\documentclass{article}

    \PassOptionsToPackage{numbers, compress}{natbib}

\usepackage[final]{neurips_2024}

\usepackage[utf8]{inputenc} 
\usepackage[T1]{fontenc}    
\usepackage{hyperref}       
\usepackage{url}            
\usepackage{booktabs}       
\usepackage{amsfonts}       
\usepackage{nicefrac}       
\usepackage{microtype}      
\usepackage{xcolor}         
\usepackage{amsmath}
\usepackage{graphicx}
\usepackage{subcaption}
\usepackage{subcaption}
\usepackage{booktabs}
\usepackage{multirow}
\usepackage{multicol}
\usepackage{makecell}
\usepackage{subcaption}
\usepackage{wrapfig}
\usepackage[inkscapelatex=false]{svg}
\usepackage{pifont}
\usepackage{tikz}
\usepackage{adjustbox}
\usepackage{cleveref}
\usepackage{float}
\usepackage{colortbl}

\newcommand{\Paragraph}[1]{\vspace{0.6mm} \noindent \textbf{#1} \hspace{0mm}}
\newcommand{\Section}[1]{\vspace{-1mm} \section{#1} \vspace{-1mm}}
\newcommand{\SubSection}[1]{\vspace{-1mm} \subsection{#1} \vspace{-1mm}}

\newcommand{\cmark}{\ding{52}}%

\usepackage{hyperref}
\hypersetup{breaklinks=true,colorlinks,citecolor=cyan}

\title{On Learning Multi-Modal Forgery Representation for Diffusion Generated Video Detection}

\author{
  Xiufeng Song$^{1}$, ~Xiao Guo$^{2}$, 
  ~Jiache Zhang$^{1}$, ~Qirui Li$^{1}$, \\ \textbf{Lei Bai$^{3}$, ~Xiaoming Liu$^{2}$, ~Guangtao Zhai$^{1}$, ~Xiaohong Liu$^{*1}$} \\
  $^1$Shanghai Jiao Tong University 
  $^2$Michigan State University 
  $^3$Shanghai Artificial Intelligence Laboratory\\
  \texttt{\{akikaze, zjc\underline{~}he, iapple1, zhaiguangtao, xiaohongliu\}@sjtu.edu.cn} \\
  \texttt{\{guoxia11, liuxm\}@cse.msu.edu} ~\texttt{baisanshi@gmail.com}\\
  $^*$ Corresponding Author
}

\begin{document}

\maketitle

\begin{abstract}
Large numbers of synthesized videos from diffusion models pose threats to information security and authenticity, leading to an increasing demand for generated content detection.
However, existing video-level detection algorithms primarily focus on detecting facial forgeries and often fail to identify diffusion-generated content with a diverse range of semantics.
To advance the field of video forensics, we propose an innovative algorithm named Multi-Modal Detection(MM-Det) for detecting diffusion-generated videos. 
MM-Det utilizes the profound perceptual and comprehensive abilities of Large Multi-modal Models (LMMs) by generating a Multi-Modal Forgery Representation (MMFR) from LMM's multi-modal space, enhancing its ability to detect unseen forgery content. 
Besides, MM-Det leverages an In-and-Across Frame Attention (IAFA)  mechanism for feature augmentation in the spatio-temporal domain. 
A dynamic fusion strategy helps refine forgery representations for the fusion.
Moreover, we construct a comprehensive diffusion video dataset, called Diffusion Video Forensics (DVF), across a wide range of forgery videos. MM-Det achieves state-of-the-art performance in DVF, demonstrating the effectiveness of our algorithm.
Both source code and DVF are available at \href{https://github.com/SparkleXFantasy/MM-Det}{link}.
\end{abstract}
\vspace{-3mm}

\Section{Introduction}
\label{sec:intro}

\begin{wrapfigure}{r}{0.45\textwidth}
    \vspace{-5mm}
    \centering
    \includegraphics[width=0.4\textwidth]{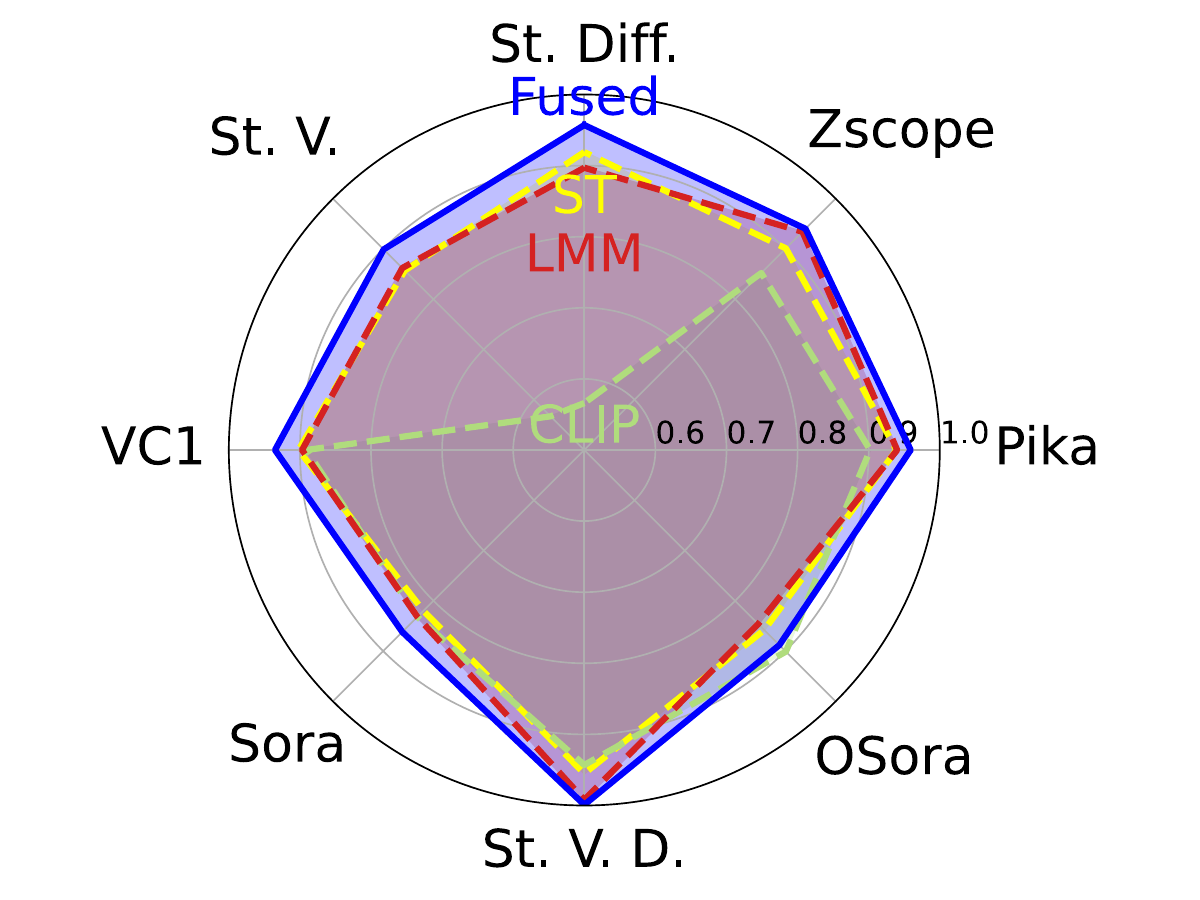}
    \caption{
    Multi-Modal Detection (MM-Det) leverages features from spatiotemporal (ST) information~(\textcolor[RGB]{236,236,8}{\rule{0.2cm}{0.24cm}}), a CLIP encoder~\cite{radford2021learning}~(\textcolor[RGB]{150,240,20}{\rule{0.2cm}{0.24cm}}), and an LMM~(\textcolor{red}{\rule{0.2cm}{0.24cm}}). 
    The Fused feature~(\textcolor{blue}{\rule{0.2cm}{0.24cm}}) achieves state-of-the-art performance in our Diffusion Video Forensics (DVF) dataset.}
    \label{fig:intro_radar}
\end{wrapfigure}

Recent years have witnessed significant advancements in diffusion generative methods, which have led to the creation of extraordinarily visually compelling content in video generation~\cite{chen2023videocrafter1, blattmann2023stable, xing2023dynamicrafter}. 
Although the latest generated videos impress society with their versatility and stability, synthetic media also poses a risk of malicious attacks, such as counterfeit faces created by deepfakes~\cite{thies2016face2face} and falsifications in business, raising public concerns about information security and privacy. 
In response to such issues, researchers have made significant progress in forgery detection, addressing problems on image editing manipulation~\cite{wu2019mantra,liu2022pscc,xiao_hifi_net++} and CNN-synthesized images~\cite{wang2020cnn,ricker2022towards,wang2023dire,ojha2023towards,guo2023hierarchical}. 
To enhance the trustworthiness and reliability of current detectors in the face of evolving generative video methods, we aim to develop a generalizable detection method for diffusion-based generative videos.

\clearpage

\begin{figure}[t]
    \centering
    \includegraphics[width=0.9\textwidth]{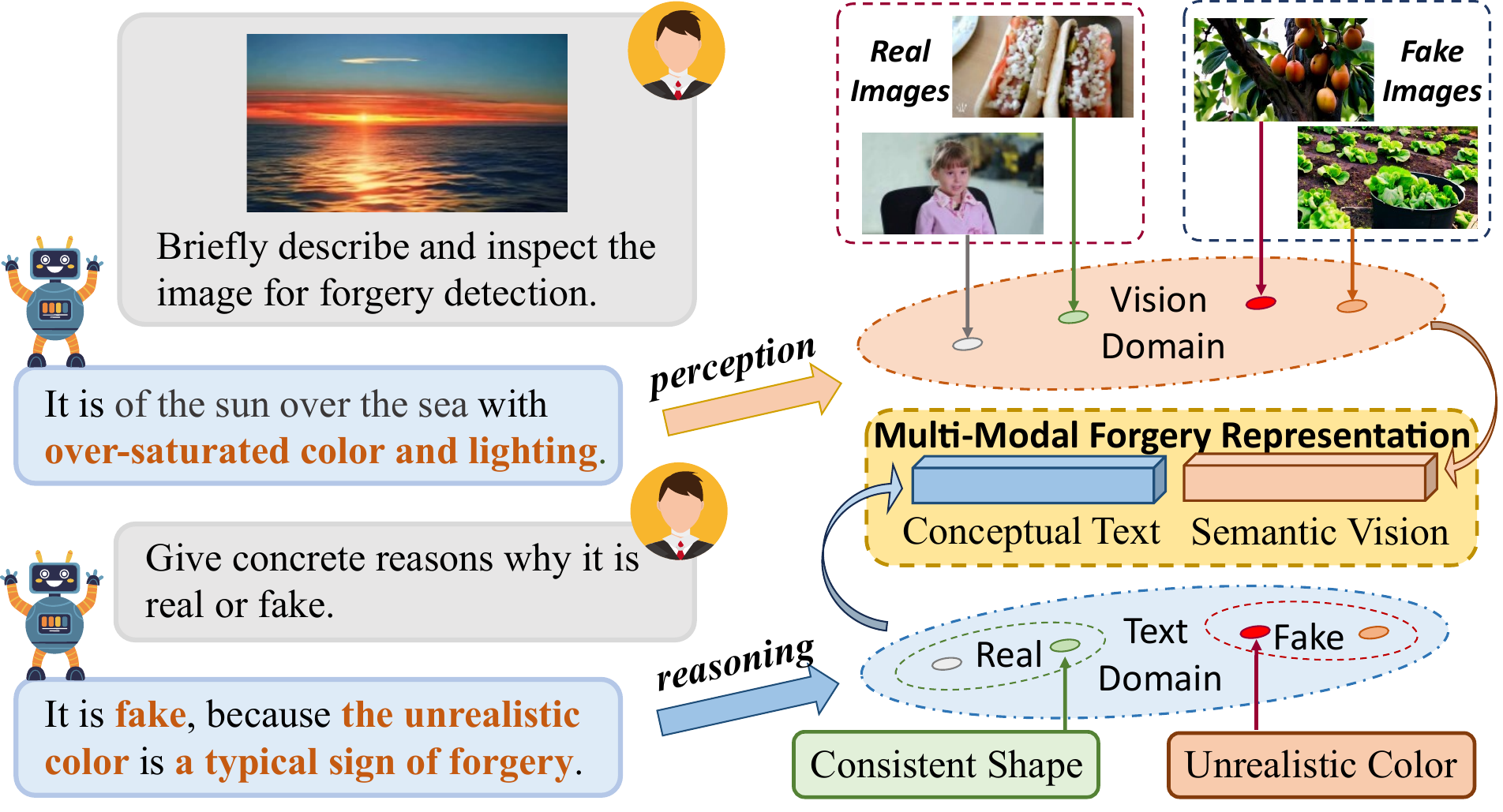}
    \caption{
    LMMs detect visual artifacts and anomalies, offering detailed textual reasoning that explains whether the image is generated using Artificial Intelligence (AI) techniques. 
    The powerful representation in the visual domain enables LMMs to understand complex contexts within frames. 
    Furthermore, their advanced language reasoning capability implicitly reveals image authenticity and provenance. 
    For instance, the term like \texttt{``consistent shape''} refers to common features in authentic content, while \texttt{``unrealistic color''} signifies typical artifacts in forged content.
    This linguistic proficiency stems from the superior perception and comprehension abilities of LMMs, contributing to a generalizable multimodal feature space.
    By leveraging the visual understanding and textual reasoning abilities of LMMs, we construct a Multi-Modal Forgery Representation (MMFR). \vspace{-4mm}
    }
    \label{fig:overview_of_main_method}
\end{figure}

Previously, the video forensics community emphasized more on developing facial forgery detection algorithms~\cite{wodajo2021deepfake,zhao2021multi,cozzolino2021id,xu2023tall,zhang2024common}, which may struggle to address recent fraudulent videos (\textit{e.g.}, sora, pika, etc.). 
Compared to facial forgery, diffusion-based generated content contains more diverse semantics, making it more challenging to distinguish diffusion forgery contents from real ones.
Towards these challenges, a new thread of research on CNN-generated image detection has emerged~\cite{wang2020cnn,ricker2022towards,wang2023dire,ojha2023towards,guo2023hierarchical}.
These works aim to learn common generation traces in image-level content, but do not design specific mechanisms to capture temporal inconsistencies in videos.

Therefore, previous defensive efforts might not be able to provide a video-level detection algorithm for newly emerged generated videos with diverse manipulation artifacts and visual contexts. 
Meanwhile, Large Multi-modal Models (LMMs) show unparalleled problem-solving ability~\cite{alayrac2022flamingo, li2022blip, li2023blip, zhu2023minigpt, zhang2024llama, liu2024visual}, thanks to its powerful multi-modal representations.
However, such representations are barely studied in the video forensics task.

Motivated by the limitation of previous work and the unprecedented understanding ability of LMMs, we propose a video-level detection algorithm, named Multi-Modal Detection (MM-Det), to capture forgery traces based on an LMM-based multi-modal representation.
MM-Det takes advantage of the perception and reasoning capabilities of LMMs to learn a generalizable forgery feature, as depicted in Fig.~\ref{fig:overview_of_main_method}. 
To the best of our knowledge, we are the first to use LMMs for video forensic work.

\begin{wrapfigure}[19]{r}{0.5\textwidth}
    \centering
    \includegraphics[width=0.48\textwidth]{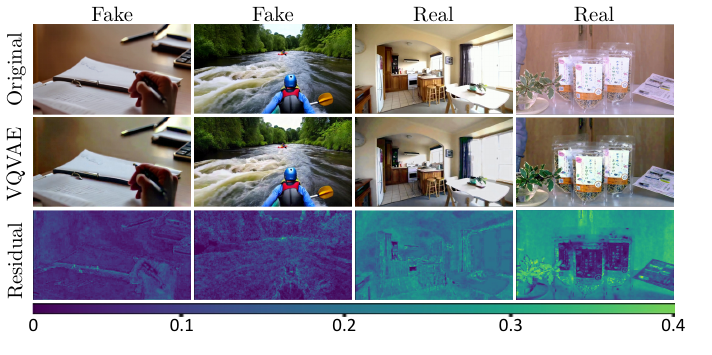}
    \caption{The residual difference between VQ-VAE~\cite{van2017neural} reconstructed images and real ones. 
    Given an encoder $\mathcal{E}$ and a decoder $\mathcal{D}$ of a VQ-VAE and taking the input video $\mathbf{v}$, the reconstructed video $\mathbf{v^{\prime}}$ is obtained as $\mathbf{v^{\prime}} = \mathcal{D}(\mathcal{E}(\mathbf{v}))$. 
    The VQ-VAE reconstruction of real images exhibits obvious edges and visible traces, whereas diffusion-generated ones are reconstructed more effectively, offering residual difference images with fewer visible traces.}
    \label{fig:recons_display}
    \vspace{1mm}
\end{wrapfigure}

Aside from multi-modal representations, two common sources of generative errors that can be leveraged for video discrimination are spatial artifacts and temporal inconsistencies. 
Our approach aims to effectively identify these two types of errors as an auxiliary feature in forgery detection. 
Inspired by the previous work~\cite{wang2023dire, ma2023exposing, ricker2024aeroblade} that shows the effectiveness of reconstruction for detecting diffusion images, we extend this idea into the video domain, amplifying diffusion artifacts both in spatial and temporal information. 
To capture such artifacts efficiently, we leverage a Vector Quantised-Variational AutoEncoder (VQ-VAE)~\cite{van2017neural} for a fast reconstruction process, as detailed in Fig.~\ref{fig:recons_display}.
Moreover, we design a novel \textbf{I}n-and-\textbf{A}cross \textbf{F}rame \textbf{A}ttention (IAFA) into a Transformer-based network, which balances frame-level forgery traces with information flow across frames, thus aggregating local and global features.

Although diffusion methods demonstrate strong capabilities in video generation, the lack of public datasets on diffusion videos hinders research efforts in the video forensic community. 
In light of this, we have established a comprehensive dataset for diffusion-generated videos, named Diffusion Video Forensics (DVF). 
DVF includes generated content from a variety of diffusion models, featuring rich semantics and high quality, serving as a general benchmark for open-world video forensics tasks.
The main contributions of this paper are as follows:

$\diamond$ We propose a detection method called MM-Det that leverages a \textbf{M}ulti-\textbf{M}odal Forgery Representation from LMMs to effectively detect diffusion-generated videos with strong generalization capability.

$\diamond$ A powerful and innovative In-and-Across Frame Attention (IAFA) mechanism is introduced to aggregate global and local patterns within forged videos, enhancing the detection of spatial artifacts and temporal inconsistencies.

$\diamond$ We introduce a large-scale dataset, named the Diffusion Video Forensics (DVF) dataset, comprising high-quality forged videos generated using $8$ diffusion-based methods. 
The DVF dataset contains diverse forgery types across videos of varying resolutions and durations, effectively serving as a benchmark for forgery detection in real-world scenarios.

$\diamond$ Our MM-Det achieves state-of-the-art detection performance on the DVF dataset. Also, a detailed analysis is provided to showcase the effectiveness of multi-modal representations in detecting forgeries, paving the way for compelling opportunities for using LMMs in future multi-media forensic research. 

\Section{Related Works}
\Paragraph{Frame-level Detector} Early work~\cite{wang2020cnn,jeong2022bihpf,gragnaniello2021gan,tan2023learning,yao2024reverse,xiao_neurips} observed that forgery traces exist in images generated by AI techniques, and such traces are commonly used as evidence to distinguish diffusion-generated content~\cite{ricker2022towards,corvi2023intriguing,corvi2023detection} and attribute if two images are generated by the same method~\cite{yu2019attributing,pan2024towards}. 
However, identifying unseen and diverse frequency-based clues in real-world scenarios is challenging. For that, existing frame-level forgery detectors concentrate on improving the generalization ability. 
For example, some works~\cite{ojha2023towards,cozzolino2023raising,xiao_hifi_net++,liu2024cfpl,liu2024forgery} introduced features from pre-trained CLIP~\cite{radford2021learning} encoders for the forensic task to help reduce the overfitting issue on specific forgery types and increase the robustness towards detection~\cite{ojha2023towards,cozzolino2023raising} and localization~\cite{xiao_hifi_net++}.
\cite{yang2024diffstega} and \cite{fu2024rawiw} proposed proactive methods to protect images from manipulation based on image watermarking and steganography.
Also, reconstruction errors through the inversion process of DDIM~\cite{song2020denoising} are studied by prior works~\cite{wang2023dire,ma2023exposing,ricker2024aeroblade,luo2024lare}
for diffusion generative content detection. 
Moreover, the previous work~\cite{yan2024transcending,tan2024rethinking,mandelli2022detecting, guo2023hierarchical} develops specific techniques that increase the generalization to unseen forgeries. 
For example, HiFi-Net~\cite{guo2023hierarchical} proposes a tree structure to model the inherent hierarchical correlation among different forgery methods,
NPR~\cite{tan2024rethinking} devises a special representation as generative artifacts, and 
the training set diversity can also contribute to generalization~ability~\cite{mandelli2022detecting}.
Unlike prior works, our MM-Det leverages multi-modal reasoning to achieve a high level of generalization ability.

\Paragraph{Video-level Detector} Early video-level methods primarily focused on detecting facial forgery. For example, 
\cite{li2020face} learned the boundary artifacts between the original background and manipulated faces. 
\cite{haliassos2021lips} discriminated fake videos from the inconsistency of mouth motion. 
\cite{zhao2021multi} designed a multi-attentional detector to capture deepfake features and artifacts. 
F3Net~\cite{qian2020thinking} captured global and local forgery traces in the frequency domain. 
\cite{cozzolino2021id,gu2021spatiotemporal,zheng2021exploring,wang2023altfreezing} explored temporal information and inconsistency from fake videos.
Most recently, DD-VQA~\cite{zhang2024common} formulates deepfake detection as a sentence-generation problem, largely improving the interpretation of deepfake detection. 
However, these studies are restricted to facial forgery methods, which are insufficient for the current defensive systems that address diverse content produced by diffusion models. 
Therefore, we develop MM-Det to detect diffusion video content, pushing forward the frontier of forgery video detection.

\Paragraph{Large Multi-modal Models (LMMs)} LMMs possess generalizable problem-solving abilities in real-world tasks, including object detection\cite{gu2021open}, semantic segmentation\cite{zhang2023next} and visual question answering\cite{wu2023next}. 
\cite{alayrac2022flamingo, li2023blip, liu2023improved, liu2024visual} studied feature alignment schemes to bridge visual and textual domains for LMMs. \cite{zhang2023video, zhang2024spartun3d, zhang2024vision} extended the boundaries of LMMs to multi-modal downstream tasks. ~\cite{su2023pandagpt} aligns multi-modal features for capabilities on cross-domain behaviors. ~\cite{wu2023cheap} developed a Large Language Model (LLM)-based feature extractor for cheap-fake detection.
Inspired by these studies, we stimulate the powerful perceptual and reasoning ability of an LMM by introducing the multi-modal feature space in video forgery detection.

\begin{figure}[tb]
  \centering
  \includegraphics[width=\linewidth]{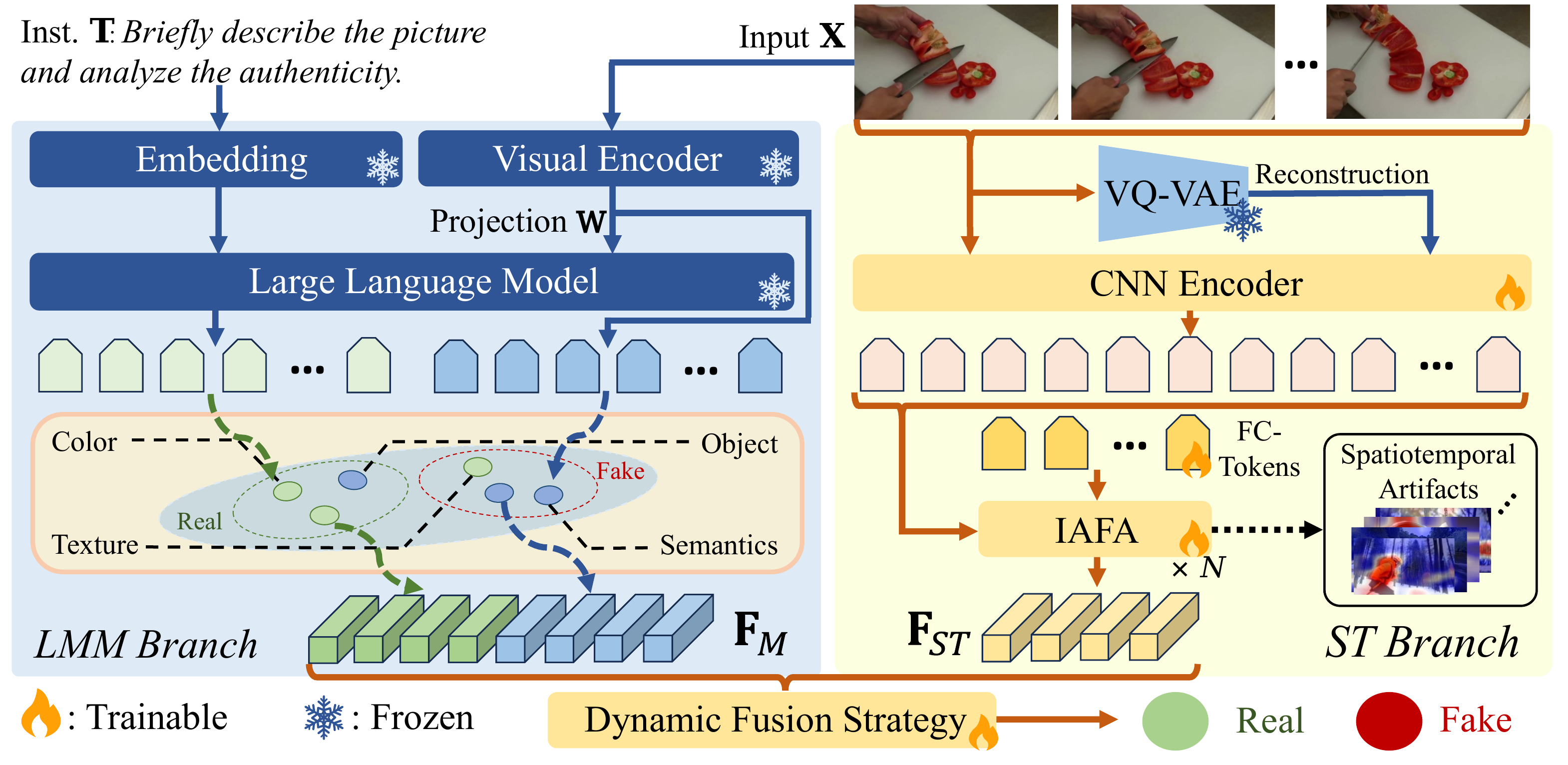}
  \caption{
  Multi-Modal Detection network (MM-Det) architecture. 
  Given an input video $\mathbf{X}$, the Large Multi-modal Model (LMM) branch takes the frame and instructions to generate Multi-Modal Forgery Representation (MMFR). 
  Hidden states from the visual encoder and large language model are extracted to form the MMFR, denoted as $\mathbf{F}_{M}$, which helps capture the forgery traces among different diffusion-generated videos. 
  In the Spatio-Temporal (ST) branch, videos are first reconstructed via a VQ-VAE, and then fed into a CNN encoder, followed by In-and-Across Frame Attention (IAFA) modules detailed in Sec. ~\ref{sec:st_forgery}. IAFA is introduced to capture features based on spatial artifacts and temporal inconsistencies, termed as $\mathbf{F}_{ST}$.
  At last, a dynamic fusion strategy combines $\mathbf{F}_{M}$ and $\mathbf{F}_{ST}$ for the final forgery prediction. 
  }
  \label{fig:overview_of_main_model}
\end{figure}

\Section{Methods}
In this section, we introduce the Multi-Modal Detection (MM-Det) framework for diffusion video detection, as depicted in Fig.~\ref{fig:overview_of_main_model}. 
More formally, Sec.~\ref{sec:mm_forgery} details a Large Multi-modal Model (LMM) branch that learns a Multi-Modal Forgery Representation (MMFR). 
Then Sec.~\ref{sec:st_forgery} reports a Spatio-Temporal (ST) branch that utilizes In-and-Across Frame Attention (IAFA) to capture spatial artifacts and temporal inconsistencies in forged videos. 
Lastly, a dynamic fusion technique reported in Sec.~\ref{sec:dynamic_fusion} adaptively combines outputs from the LMM branch and ST branch. 

\SubSection{Multi-Modal Forgery Representation}
\label{sec:mm_forgery} 
We propose a novel Multi-Modal Forgery Representation (MMFR) from the multi-modal space of LMMs in LMM branch. This representation utilizes the powerful perceptual and reasoning abilities of LMMs in the form of instruction-based conversations.

Specifically, LMM branch is built on the top of LLaVA~\cite{liu2024visual}, one representative LMM, which has two key components: a visual encoder (\textit{e.g.}, $\mathcal{D}_v$), instantiated by visual encoders from the Contrastive Language-Image Pre-Training (CLIP)~\cite{radford2021learning}, and the large language model $\mathcal{D}_L$ (\textit{i.e.}, Llama 2~\cite{touvron2023llama}). 
Let us denote the input video as $\mathbf{X} \in \mathbb{R}^{N \times H \times W \times C}$ that contains $N$ frames, where each frame is represented as $\mathbf{x} \in \mathbb{R}^{H \times W \times C}$.
First, $\mathbf{x}$ is fed to $\mathcal{D}_v$ to obtain the corresponding visual representation $\mathbf{F}_{V} \in \mathbb{R}^{Z}$. 
Such $\mathbf{F}_{V}$ not only contains rich semantics but also shows impressive generalization ability and robustness in the forgery detection task~\cite{ojha2023towards,sha2023fake,cozzolino2024raising}.
Then, a textual instruction $\mathbf{T}$ is sampled from pre-defined templates $\mathbf{Q}$ to guide the LMM on forgery detection reasoning. 
Both visual representation $\mathbf{F}_{V}$ and instruction $\mathbf{T}$ are fed to $\mathcal{D}_L$, which generates enhanced visual representations (\textit{e.g.}, $\mathbf{F}_{L}$).
We convert $\mathbf{F}_{V}$ into a sequence of visual tokens~\cite{liu2024visual} (\textit{i.e.}, $\mathbf{H}_{v}=\{\mathbf{h}_{v,m}\}^{M}_{m=1} \in \mathbb{R}^{M\times D}$), and $\mathbf{T}$ is transformed into textual tokens $\mathbf{H}_{t}=\{\mathbf{h}_{t,o}\}^{O}_{o=1} \in \mathbb{R}^{O\times D}$. 
Both $\mathbf{H}_{v}$ and $\mathbf{H}_{t}$ are taken as the input to $\mathcal{D}_L$, generating $\mathbf{F}_{L} \in \mathbb{R}^{S\times D}$ that can be tokenized into the language response providing reasoning (Fig.~\ref{fig:overview_of_main_method}) about the authenticity of the input $\mathbf{x}$.
This procedure is formulated as 
\begin{equation}
    \mathbf{F}_L = \mathcal{D}_L(\mathbf{H}_{t}, \mathbf{H}_{v}) = \mathcal{D}_L(\mathbf{T}, \mathbf{F}_{V}),
\end{equation}
where $\mathbf{T}$ guides the pre-trained $\mathcal{D}_L$ in comprehending visual content (\textit{i.e.}, $\mathbf{F}_{V}$), discerning the subset information from $\mathbf{F}_V$. 
This instruction $\mathbf{T}$ enables LMM branch to obtain the multi-modal representation that leverages the generalization ability from the pre-trained large language model Llama 2 (\textit{i.e.}, $\mathbf{D}_{L}$), being different to prior work~\cite{ojha2023towards,cozzolino2024raising} that only relies on $\mathbf{F}_{E}$.

Lastly, we retrieve the final MMFR, denoted as $\mathbf{F}_M \in \mathbb{R}^{M \times Z}$, by concatenating $\mathbf{F}_{V}$ and $\mathbf{F}_{L}$ after a linear layer(\textit{i.e.}, \texttt{PROJ}), as 
\begin{equation}
    \mathbf{F}_M = \texttt{CONCAT}(\{\texttt{PROJ}(\mathbf{F}_{V}), \texttt{PROJ}(\mathbf{F}_{L})\}).
\end{equation}

\begin{figure}[tb]
  \begin{subfigure}{0.55\textwidth}
  \includegraphics[width=\textwidth]{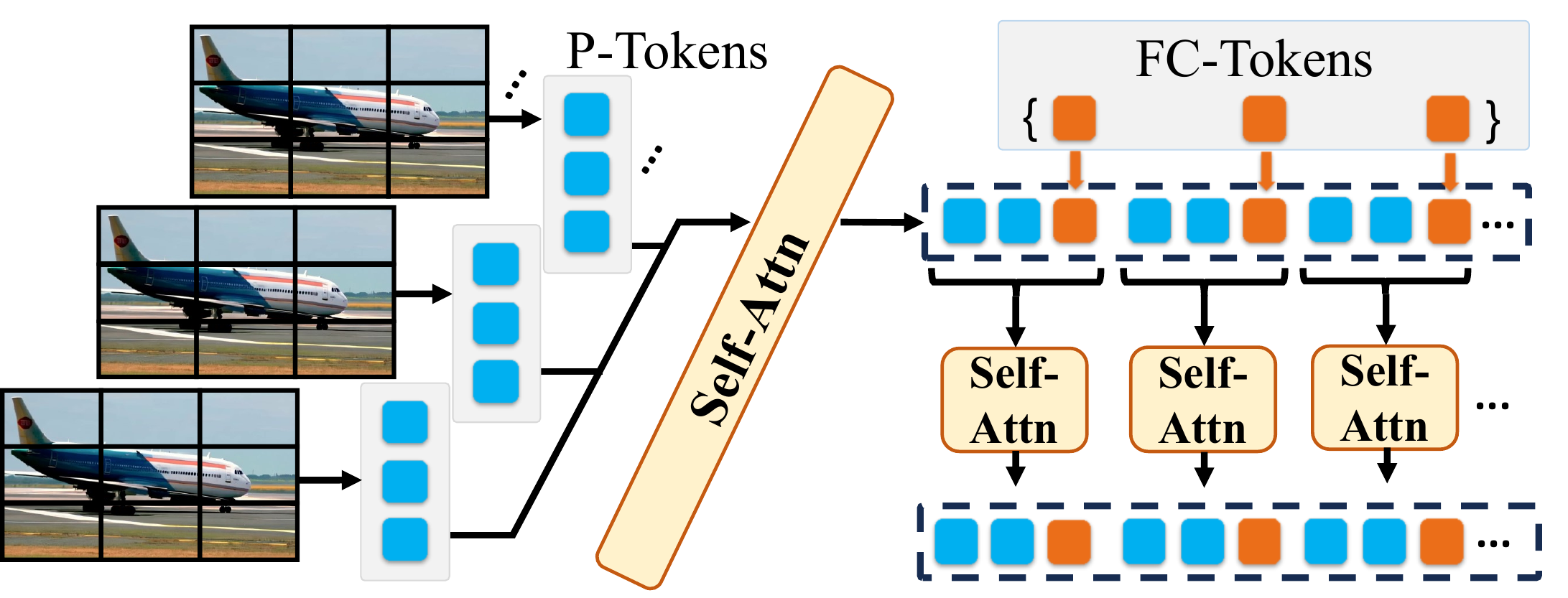}
    \caption{Mechanism of In-and-Across Frame Attention}
    \label{fig:method_fc_attn}
  \end{subfigure}
  \begin{subfigure}{0.42\textwidth}
  \includegraphics[width=1.\textwidth]{figs/pdfs/method_dynamic_fusion.pdf}
    \caption{Dynamic Fusion Strategy}
  \label{fig:method_dynamic_fusion}

  \end{subfigure}
  \caption{
  (a) In-and-Across Frame Attention (IAFA): Each input frame (or its feature map) is divided into patches that are transformed into tokens, termed P-tokens (\textcolor{cyan}{\rule{0.3cm}{0.25cm}}).
  We introduce additional frame-centric tokens FC-tokens (\textcolor{orange}{\rule{0.3cm}{0.25cm}}) encapsulating the global forgery information of the video frame. 
  In each transformer layer, self-attention is applied alternately among all P-tokens from video frames as well as among the same frame’s P-tokens and its FC-tokens. 
  (b) The dynamic fusion strategy captures that takes $\mathbf{F}_{ST}$ and $\mathbf{F}_{M}$ as inputs and output channel-wise dependencies, which help refine forgery representations for the fusion.
  }
\end{figure}

\SubSection{Capturing Spatial-Temporal Forgery Traces}
\label{sec:st_forgery}

Targeting capturing spatiotemporal artifacts in video tasks, we introduce a Spatial-Temporal (ST) branch that learns effective diffusion forgery representation at the video level.
Through a reconstruction procedure, we amplify the diffusion traces in the frequency domain, which is then captured by In-and-Across Frame Attention (IAFA) to form an effective video-level feature.

\Paragraph{Amplification of Diffusion Traces}
Similar to prior studies ~\cite{wang2023dire,ma2023exposing} that discovered specific generative traces of diffusion models through reconstruction on diffusion-generated images, we utilize an Autoencoder to amplify diffusion traces in videos. The reconstruction procedure can be expressed as follows.

More formally, denote the input video as $\mathbf{X} \in \mathbb{R}^{N \times H \times W \times C}$ that contains $N$ frames, in which each frame is represented as $\mathbf{x} \in \mathbb{R}^{H \times W \times C}$. 
We leverage a VQ-VAE~\cite{van2017neural} to obtain the reconstructed version of $\mathbf{x}$, which is denoted as $\mathbf{\hat{x}} \in \mathbb{R}^{H \times W \times C}$.
The difference between $\mathbf{x}$ and $\mathbf{\hat{x}}$ makes an effective indicator of showing if the input is generated by diffusion models, as depicted in Fig.~\ref{fig:recons_display}. 
Therefore, we jointly proceed $\mathbf{x}$ and $\mathbf{\hat{x}}$ to the following proposed modules
for learning an effective representation of discerning forgeries. 

It is worth mentioning that the prior approach~\cite{wang2023dire} also adopts the idea of using the residual difference between original and reconstructed inputs to help forgery detection, but the reconstruction method requires multiple time-step denoising operations, which are computationally infeasible to reconstruct all frames from $\mathbf{X}$. 
In contrast, our VQ-VAE-based reconstruction method only requires one single forward propagation to obtain reconstructed frame $\mathbf{\hat{x}}$, meanwhile preserving the effectiveness in indicating the discrepancy between real and fake inputs.

\Paragraph{Integration of Spatial and Temporal Information} 
Same as the previous work~\cite{arnab2021vivit,neimark2021video} that utilizes ViT to learn video-level information, we transform each input video frame (\textit{i.e.}, $\mathbf{x}$) into $L$ tokens, and propose IAFA for information aggregation, as depicted in Fig.~\ref{fig:method_fc_attn}.
Let us denote frame-level tokens as Patch-wise tokens (P-tokens), as they represent local information of one patch from $\mathbf{x}$.
More formally, the $i$th frame, ${\mathbf{x}_i}$ is divided into $L$ patches, and all patches are projected into $\mathbf{T}_{i} = \{ \mathbf{t}^{j}_{i}\}^{L}_{j=1} \in \mathbb{R}^{L\times D}$, where $D$ represents the dimension of each P-token.
Also, to capture the global forgery information for each video frame, we introduce additional tokens called Frame-Centric tokens (FC-tokens).
The FC-token is denoted as $\mathbf{p}_{i} \in \mathcal{R}^{D}$ for frame ${\mathbf{x}_i}$ and attends other tokens \textit{within} the same frame $\mathbf{x}_{i}$.

During the forward propagation, we conduct IAFA based on a Transformer, with each block containing two self-attentions and consecutively modeling the local and global forgeries at each video frame. 
Specifically, the first self-attention captures dependencies among P-tokens that restore local forgery clues.
This is formulated by Eq.~\ref{eq:cross_frame_attn} that $\mathbf{t}_{i}^{j} \in \mathbb{R}^{D}$ attends the token $\mathbf{t}_{q}^{p} \in \mathbb{R}^{D}$ that represents $p$th token from $q$th frame ${\mathbf{x}_{q}}$.
Consequently, given the $i$th frame \textit{i.e.}, $\mathbf{x}_{i}$, the second self-attention is conducted among the FC-token (\textit{i.e.}, $\{\mathbf{p}_{i}\}$) and P-tokens (\textit{i.e.}, $\{\mathbf{t}_{i}^{0}$, $\mathbf{t}_{i}^{1}$, ... $\mathbf{t}_{i}^{L-1}\}$) from the same frame, which encapsulates patch-wise forgery information into the global one for learning the more robust representation.
We formulate this procedure in Eq.~\ref{eq:in_frame_attn}.
\begin{equation}
    \vspace{-2mm}
    \mathbf{t}_{i}^{j} = \sum \texttt{ATTN}(\mathbf{t}_{i}^{j}, \mathbf{t}_{q}^{p}) \quad \text{$ i,j \in [1, N]$, $j,p \in [1, L]$},
    \label{eq:cross_frame_attn}
\end{equation}
\begin{equation}
    \mathbf{p}_{i} = \sum \texttt{ATTN}(\mathbf{p}_{i}, \mathbf{t}_{i}^{j}) \quad \text{$ j \in [1, L]$},
    \label{eq:in_frame_attn}
\end{equation}
where $\texttt{ATTN}$ refers to the self-attention operation.

\SubSection{Dynamic Fusion}
\label{sec:dynamic_fusion}
We devise the dynamic fusion strategy (\textit{i.e.}, $\mathcal{D}_{f}$) that combines spatiotemporal information from ST branch and MMFR (\textit{i.e.}, $\mathbf{F}_{f}$ and $\mathbf{F}_{m}$) for the final prediction, by adjusting their contributions based on forgeries from the input. 
More formally, $\mathcal{D}_{f}$ (Fig.~\ref{fig:method_dynamic_fusion}) learns channel-wise dependencies among forgery representations (\textit{e.g.}, $\mathbf{F}_{ST}$ and $\mathbf{F}_{M}$) via the attention mechanism, generating $\mathbf{w} \in \mathbb{R}^{N+M}$ as the output.
This procedure can be expressed as $\mathbf{w} = \mathcal{D}_{f}(\texttt{CONCAT}\{\mathbf{F}_{ST}, \mathbf{F}_M\})$.
Also, $\mathbf{w}$ contains $\mathbf{w}_{ST} \in \mathbb{R}^{N}$ and $\mathbf{w}_M \in \mathbb{R}^{M}$, representing learned channel-wise weights for $\mathbf{F}_{ST}$ and $\mathbf{F}_{M}$, respectively.
Such channel-wise weights are important in the fusion purpose, as they help emphasize useful information --- we use $\mathbf{w}_{ST}$ and $\mathbf{w}_{M}$ to refine forgery representations as $\mathbf{F}^{\prime}_{ST} = \mathbf{F}_{ST} \mathbf{w}_{ST}$ and $\mathbf{F}^{\prime}_{M} = \mathbf{F}_{M} \mathbf{w}_{M}$. 
Lastly, $\mathbf{F}^{\prime}_{ST}$ and $\mathbf{F}^{\prime}_{M}$ are concatenated into the fused representation $\mathbf{F}_0 \in \mathbb{R}^{(M+N)\times D}$, which is used for the final scalar prediction $s$ via the average pooling (\textit{i.e.}, \texttt{AVG}) and linear layers(\textit{i.e.}, \texttt{PROJ}):
\begin{equation}
    s = \texttt{PROJ}(\texttt{AVG}(\mathbf{F}_0)) = \texttt{PROJ}(\texttt{AVG}(\texttt{CONCAT}\{\mathbf{F}^{\prime}_{ST}, \mathbf{F}^{\prime}_{M}\}).
\end{equation}

\begin{figure}[t]
\centering
\includegraphics[width=1.\linewidth]{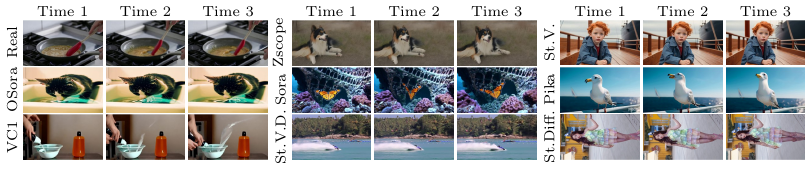}
\caption{Sampled videos from DVF dataset. DVF contains $8$ video generation methods, including $7$ text-to-video methods and $1$ image-to-video method. Real videos are selected from Internvid-$10$M~\cite{wang2023internvid} and Youtube-$8$M~\cite{abu2016youtube}. [Key: OSora: OpenSora; VC1: Videocrafter1~\cite{chen2023videocrafter1}; Zscope: Zeroscope; St. V. D.: Stable Video Diffusion~\cite{blattmann2023stable}; St.Diff.: Stable Diffusion~\cite{rombach2022high}; St. V.:Stable Video]}\vspace{-3mm}
\label{fig:dvf_dataset_samples}
\end{figure}

\Section{Diffusion Video Forensics (DVF) Dataset}
We construct a large-scale dataset for the video forensic task named Diffusion Video Forensics (DVF), as shown in Fig. ~\ref{fig:dvf_dataset_samples}. 
DVF contains $8$ diffusion generative methods, including Stable Diffusion~\cite{rombach2022high}, VideoCrafter$1$~\cite{chen2023videocrafter1}, Zeroscope, Sora, Pika, OpenSora, Stable Video, and Stable Video Diffusion\cite{blattmann2023stable}. 

\begin{figure}[t]
\centering
\captionsetup[subfloat]{labelfont=footnotesize,textfont=footnotesize}
\renewcommand{\thesubfigure}{{a}}
\subfloat[Generation pipeline]{\includegraphics[width=0.41\linewidth]{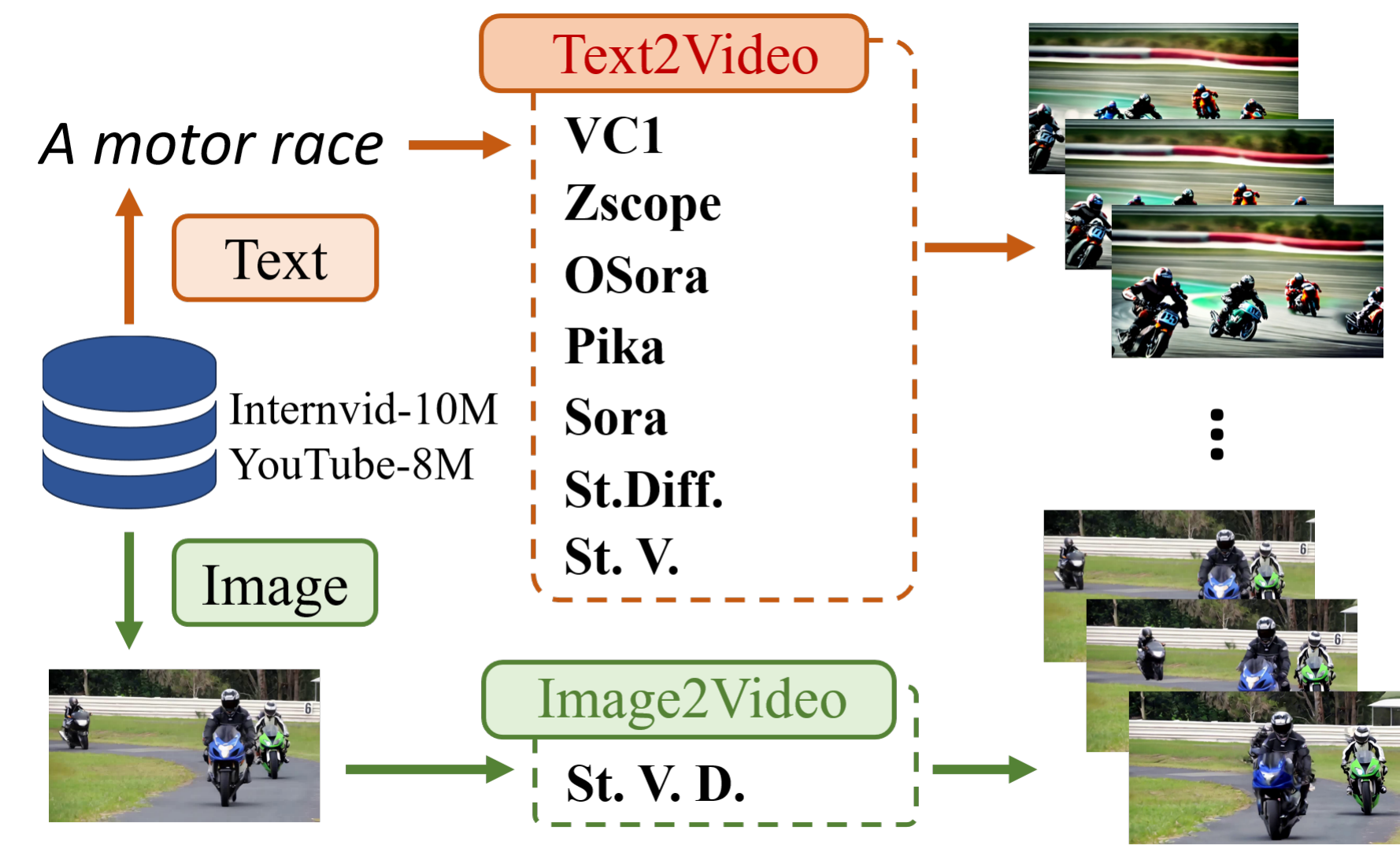}}
\renewcommand{\thesubfigure}{{b}}
\subfloat[Durations and resolutions]{\includegraphics[width=0.26\linewidth]{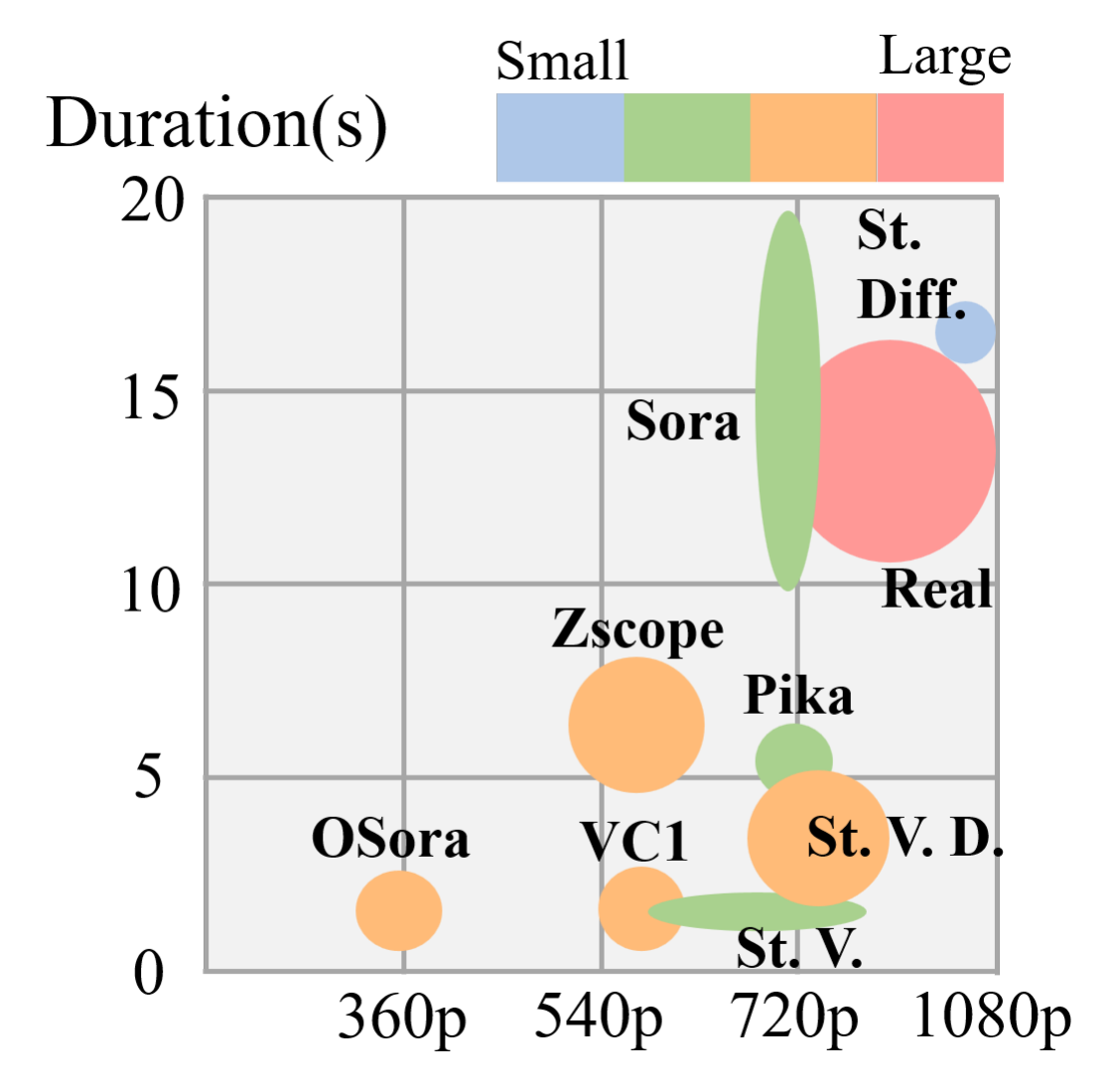}}
\renewcommand{\thesubfigure}{{c}}
\subfloat[Statistics of DVF]{\includegraphics[width=0.3\linewidth]{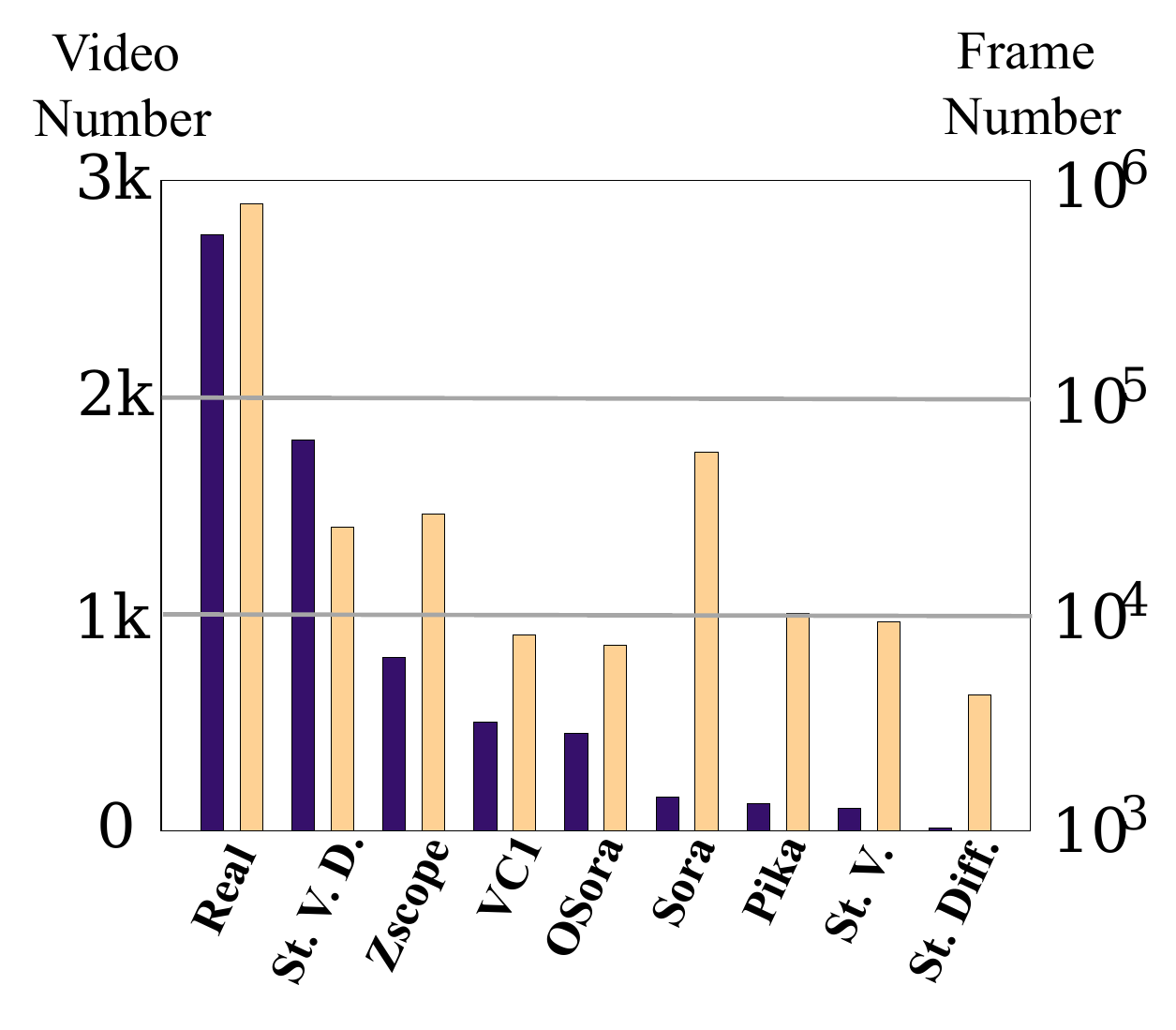}}
\caption{The overview of DVF dataset:
(a) The procedure of forged video generation and collection. Real frames and captions sampled from Internvid-10M~\cite{wang2023internvid} and Youtube-8M~\cite{abu2016youtube} for text-to-video and image-to-video generation. 
(b) DVF contains videos at various resolutions and durations. (c) The statistics of DVF, measured by the numbers of frames and videos.[Key: VC1: Videocrafter1~\cite{chen2023videocrafter1}; Zscope: Zeroscope; OSora: OpenSora; St.Diff.: Stable Diffusion~\cite{rombach2022high}; St. V.:Stable Video; St. V. D.: Stable Video Diffusion~\cite{blattmann2023stable}]
}
\label{fig:overview_dvf}
\end{figure}

To efficiently streamline the collection, we construct an effective automated pipeline that generates forgery videos based on real videos and prompts. 
Specifically, we start from two real video datasets, Internvid-$10$M~\cite{wang2023internvid} and Youtube-$8$M~\cite{abu2016youtube}. 
Real videos are sampled for rich semantic content, with their frames and captions used for generation. Fig.~\ref{fig:overview_dvf}~\textcolor{red}{a} introduces the generation process of DVF. 
For open-sourced generation methods, a prompt is fed to a text-to-video method(\textit{i.e.} VideoCrafter$1$, Zeroscope, OpenSora), or a frame is provided to an image-to-video method(Stable Video Diffusion) to generate the corresponding fake video.
For commercial and close-sourced datasets(\textit{i.e.} Stable Diffusion, Stable Video, Sora, Pika), forgery videos are collected from official websites and social media. 
In total, we collect $3,938$ fake videos and $2,750$ real videos in DVF. As shown in Fig.~\ref{fig:overview_dvf}~\textcolor{red}{b} and Fig.~\ref{fig:overview_dvf}~\textcolor{red}{c}, our dataset contains multiple resolutions and durations. 
The video number of each dataset varies from $0.1$k to $2.8$k, with the corresponding frame numbers from $4.2$k to $784$k. 
More details about DVF are provided in Appendix~\ref{sec:supple_dvf}.

\Section{Training Strategy}
\label{sec:train_strategy}

This section details our two-stage training strategy, in which we first finetune the LMM branch via instruction tuning~\cite{liu2024visual} and then optimize the entire framework in an end-to-end manner.

\paragraph{LMM Branch Instruction Tuning} 
We first adapt LLaVA~\cite{liu2024visual} to the forgery detection downstream task based on instruction tuning, an empirically effective way for various downstream tasks, which leverages LoRA~\cite{hu2021lora} to improve the reasoning ability of Large Language Models(LLMs). 
For that, construct a large image-text paired dataset, named Rich Forgery Reasoning Dataset. 
Please refer to Appendix ~\ref{sec:supple_rfrd} for more details. 
We use multi-turn conversations to fine-tune the LMM, enhancing its ability to identify and judge the authenticity of input images. 
Following the instruction tuning strategy of LLaVA~\cite{liu2024visual}, we only fine-tune the projection layers and LLM in LLaVA.
More formally, we formulate the objective function as the loss for an auto-regressive model, which is based on answer tokens from the LLM, as:
\begin{equation}
    \vspace{-2mm}
    \mathcal{L}(\theta_1) = - \sum_{t=1}^T log(p_{\theta_1}(s^t | s^{i < t})),
\end{equation}
where $s^i$ refers to the $i^{th}$ prediction token, $T$ refers to the length of total prediction tokens, and $\theta_1$ refers to the trainable parameters in the LMM. 

\paragraph{End-to-End Training}
After fine-tuning LLaVA, we use this model to form LMM branch of MM-Det, and then the entire model is trained in an end-to-end manner.
Please note that all parameters in LMM branch are frozen to ensure the optimal multi-modal representation can be obtained. 
More formally, we denote MM-Det's final prediction scalar and the ground truth as $v$ and $y$, respectively, and the model is optimized by the cross-entropy loss $\mathcal{L}$ as follows:

\begin{equation}
    \mathcal{L}(\theta_2) = -{(y\log \mathcal{D}(v) + (1 - y)\log(1 - \mathcal{D}(v}))
\end{equation}

where $\theta_2$ refers to trainable parameters in both ST branch and dynamic fusion modules.

\Section{Experiments}
\SubSection{Setup}
In the experiment, we use the proposed DVF for the evaluation.
In training, $1,000$ videos from YouTube and $1,973$ fake videos generated by Stable Video Diffusion serve as the training set, in which $90\%$ are used for training and the remaining $10\%$ for validation. 
Real videos from Internvid-$10$M~\cite{wang2023internvid} and fake videos from $6$ generative methods are used as testing samples.
More details on training and testing are provided in Appendix~\ref{sec:supple_implementation}.

For a fair comparison, we choose the following $10$ recent detection methods as
baselines. 
CNNDet~\cite{wang2020cnn} applies
a ResNet~\cite{he2016deep} as the backbone for forgery detection. F3Net~\cite{qian2020thinking} utilizes frequency traces left in forgery content. 
HiFi-Net~\cite{guo2023hierarchical} devise a specific hierarchical fine-grained learning scheme to learn a wide range of forgery traces.
Clip-Raising~\cite{cozzolino2024raising}, Uni-FD~\cite{ojha2023towards} takes advantage of a pre-trained CLIP~\cite{radford2021learning} as a training-free feature space. 
DIRE~\cite{wang2023dire} detects diffusion images based on a reconstruction process of DDIM~\cite{song2020denoising}. 
ViViT~\cite{arnab2021vivit}, TALL~\cite{xu2023tall}, and TS2-Net~\cite{liu2022ts2} take advantage of spatiotemporal information in various visual tasks. 
DE-FAKE~\cite{sha2023fake} adopts visual and textual representations
based on a CLIP encoder for image forgery detection. 
For the measurement, we choose AUC since it is a threshold-independent metric. We computer means and deviations of the performance across 5 runs on different random seeds.

\begin{table}[tb]
  \caption{Video forgery detection performance on the DVF dataset measured by AUC ($\%$). [Key: \textbf{\textcolor{red}{Best}}; \textcolor{blue}{Second Best}; Stable Diff.: Stable Diffusion; Avg.: Average]}
  \vspace{1mm}
  \label{tab:expt_cross_dataset_evaluation}
  \centering
  \resizebox{1\linewidth}{!}{

  \begin{tabular}{c|ccccccc|c}
     \hline
    \multirow{2}{*}{Method} &  \multirow{2}{*}{{\makecell{Video-\\Crafter$1$}}}  & \multirow{2}{*}{{\makecell{Zero-\\scope}}} & \multirow{2}{*}{{\makecell{Open-\\Sora}}} & \multirow{2}{*}{{\makecell{Sora}}} & \multirow{2}{*}{{\makecell{Pika}}} & \multirow{2}{*}{{\makecell{Stable \\ Diff.}}} & \multirow{2}{*}{{\makecell{Stable \\ Video}}} & \multirow{2}{*}{{\makecell{Avg.}}} \\
    & & & & & & & & \\
    \hline
    \rowcolor{gray!30}CNNDet~\cite{wang2020cnn}  & $87.4 {\scriptstyle \,\pm\, 1.5}$ & $88.2 {\scriptstyle \,\pm\, 1.5}$ & $78.0 {\scriptstyle \,\pm\, 1.6}$ & $63.8 {\scriptstyle \,\pm\, 3.5}$  & $77.3 {\scriptstyle \,\pm\, 2.1}$ & $73.5 {\scriptstyle \,\pm\, 2.4}$ & $78.9 {\scriptstyle \,\pm\, 4.1}$ & $78.2 {\scriptstyle \,\pm\, 1.3}$ \\
    DIRE~\cite{wang2023dire}  & $55.9 {\scriptstyle \,\pm\, 2.2}$ & $61.8{\scriptstyle \,\pm\, 3.3}$ & $53.8 {\scriptstyle \,\pm\, 1.8}$ & $60.5 {\scriptstyle \,\pm\, 5.5}$ & $65.8 {\scriptstyle \,\pm\, 1.7}$ & $62.7 {\scriptstyle \,\pm\, 3.6}$ & $69.9 {\scriptstyle \,\pm\, 2.5}$ & $62.1 {\scriptstyle \,\pm\, 1.8}$\\
    \rowcolor{gray!30}Raising~\cite{cozzolino2023raising}   & $63.8 {\scriptstyle \,\pm\, 1.6}$ & $60.7 {\scriptstyle \,\pm\, 1.9}$ & $64.1 {\scriptstyle \,\pm\, 1.9}$ & $68.8 {\scriptstyle \,\pm\, 3.6}$  & $70.7 {\scriptstyle \,\pm\, 1.4}$ & $78.2 {\scriptstyle \,\pm\, 2.3}$ & $62.8 {\scriptstyle \,\pm\, 1.9}$ & $67.0 {\scriptstyle \,\pm\, 0.9}$\\
    Uni-FD~\cite{ojha2023towards}  & $75.0 {\scriptstyle \,\pm\, 3.0}$ & $71.2 {\scriptstyle \,\pm\, 3.4}$ & $76.6 {\scriptstyle \,\pm\, 2.2}$ & $73.1 {\scriptstyle \,\pm\, 1.5}$  & $76.2 {\scriptstyle \,\pm\, 2.4}$ & $80.2 {\scriptstyle \,\pm\, 1.9}$ &$66.7 {\scriptstyle \,\pm\, 2.6}$ & $74.1 {\scriptstyle \,\pm\, 1.2}$\\
    \rowcolor{gray!30}F3Net~\cite{qian2020thinking} & $89.7 {\scriptstyle \,\pm\, 1.8}$ & $80.5 {\scriptstyle \,\pm\, 2.2}$ & $69.3 {\scriptstyle \,\pm\, 1.8}$ & $70.8 {\scriptstyle \,\pm\, 6.9}$ & \textcolor{blue}{$88.9 {\scriptstyle \,\pm\, 2.3}$ } & $84.4 {\scriptstyle \,\pm\, 2.1}$ & \textcolor{blue}{$85.1 {\scriptstyle \,\pm\, 1.3}$} & $81.3 {\scriptstyle \,\pm\, 1.9}$ \\
    ViViT~\cite{arnab2021vivit} & $79.1 {\scriptstyle \,\pm\, 3.1}$ & $78.4 {\scriptstyle \,\pm\, 2.0}$ & $77.7 {\scriptstyle \,\pm\, 2.3}$ &$69.4 {\scriptstyle \,\pm\, 3.5}$ & $83.1 {\scriptstyle \,\pm\, 2.6}$ & $82.1 {\scriptstyle \,\pm\, 2.0}$ & $83.6 {\scriptstyle \,\pm\, 2.1}$ & $79.1 {\scriptstyle \,\pm\, 1.8}$\\
    \rowcolor{gray!30}TALL~\cite{xu2023tall} & $76.0 {\scriptstyle \,\pm\, 1.4}$ & $65.9 {\scriptstyle \,\pm\, 1.6}$ & $62.1 {\scriptstyle \,\pm\, 1.3}$ & $64.3 {\scriptstyle \,\pm\, 1.9}$ & $72.3 {\scriptstyle \,\pm\, 2.9}$ & $65.8 {\scriptstyle \,\pm\, 2.8}$ & $79.8 {\scriptstyle \,\pm\, 2.2}$ & $69.5 {\scriptstyle \,\pm\, 1.4}$\\
    TS2-Net~\cite{liu2022ts2} & $61.8 {\scriptstyle \,\pm\, 3.9}$ & $70.6 {\scriptstyle \,\pm\, 2.8}$ & $75.5 {\scriptstyle \,\pm\, 3.4}$ & \textcolor{blue}{$78.0 {\scriptstyle \,\pm\, 2.9}$} & $78.2 {\scriptstyle \,\pm\, 2.8}$ & $62.1 {\scriptstyle \,\pm\, 3.8}$ & $78.6 {\scriptstyle \,\pm\, 3.0}$ & $72.1 {\scriptstyle \,\pm\, 2.8}$ \\
    \rowcolor{gray!30}DE-FAKE~\cite{sha2023fake} & $74.7 {\scriptstyle \,\pm\, 1.7}$ & $68.2 {\scriptstyle \,\pm\, 2.9}$ & $55.8 {\scriptstyle \,\pm\, 3.6}$ & $64.1 {\scriptstyle \,\pm\, 3.1}$ & $85.6 {\scriptstyle \,\pm\, 2.2}$ & $85.4 {\scriptstyle \,\pm\, 2.6}$ & $70.6 {\scriptstyle \,\pm\, 1.9}$ & $72.1 {\scriptstyle \,\pm\, 2.2}$ \\
    HiFi-Net~\cite{guo2023hierarchical} & \textcolor{blue}{$90.2 {\scriptstyle \,\pm\, 3.0}$} & \textcolor{blue}{$89.7 {\scriptstyle \,\pm\, 2.9}$} & \textcolor{blue}{$80.1 {\scriptstyle \,\pm\, 2.6}$} & $70.1 {\scriptstyle \,\pm\, 3.8}$ &$87.8 {\scriptstyle \,\pm\, 2.9}$ & \textcolor{blue}{$89.2 {\scriptstyle \,\pm\, 2.5}$} & $83.1 {\scriptstyle \,\pm\, 2.2}$& \textcolor{blue}{$84.3 {\scriptstyle \,\pm\, 2.4}$}\\
    \hline
    \rowcolor{gray!30}MM-Det (Ours) & \textcolor{red}{$\mathbf{93.5} {\scriptstyle \,\pm\, 3.6}$} & \textcolor{red}{$\mathbf{94.0} {\scriptstyle \,\pm\,2.8}$} & \textcolor{red}{$\mathbf{88.8} {\scriptstyle \,\pm\, 2.8}$} & \textcolor{red}{$\mathbf{86.2} {\scriptstyle \,\pm\, 1.8}$} &\textcolor{red}{$\mathbf{95.9} {\scriptstyle \,\pm\, 2.8}$} & \textcolor{red}{$\mathbf{95.7} {\scriptstyle \,\pm\, 2.5}$} & \textcolor{red}{$\mathbf{89.9} {\scriptstyle \,\pm\, 2.0}$} &  \textcolor{red}{$\mathbf{92.0} {\scriptstyle \,\pm\, 2.6}$} \\   \hline
  \end{tabular}
  }\vspace{-2mm}
\end{table}

\SubSection{Video Forgery Detection Performance}
In Tab.~\ref{tab:expt_cross_dataset_evaluation}, our proposed MM-Det achieves SoTA performance in detecting diffusion video, surpassing the second-best method, \textit{i.e.}, HiFi-Net, by $7.7\%$ in the average of AUC scores.
Specifically, for prior methods that are based on pre-trained CLIP features, such as Raising~\cite{cozzolino2023raising} and Universal FD~\cite{ojha2023towards}, they remain effective on certain types of diffusion content (\textit{i.e.} Stable Diffusion), but fail on most others. 
Simple structures like CNN~\cite{wang2020cnn} exceed these CLIP-based methods after being fine-tuned on our proposed DVF, reaching an average AUC score of $78.2\%$, which proves the necessity of such datasets. 
As for our method, MM-Det outperforms other methods in most datasets. 
Compared with frequency-based forgery methods, \textit{e.g.}, HiFi-Net~\cite{qian2020thinking} and CLIP-based methods~\cite{cozzolino2023raising, ojha2023towards}, our method improves the performance from $+3.3\%$(VideoCrafter$1$) to $+15.5\%$(Stable Diffusion). 
It is worth mentioning that HiFi-Net makes the second-best performer in our DVF dataset, achieving $84.3\%$ AUC scores. 
We believe this indicates the multi-branch feature extractor used in HiFi-Net carries versatile forgery traces at multiple resolutions, enhancing the learning of the forgery invariant.
The failure of frequency traces and CLIP features raises the need for more effective features. As for spatiotemporal baselines~\cite{arnab2021vivit, xu2023tall, liu2022ts2}, we outperform them by $+12.9\%$(ViViT), $+22.5\%$(TALL) and $+19.9\%$(TS2-Net), demonstrating the effective features of MMFR and IAFA. 
At last, our detector improves by $19.9\%$ to another multi-modal detector~\cite{wodajo2021deepfake}, which utilizes visual information and corresponding captions for feature enhancement. 
It is shown that the introduction of MMFR is more generalizable than a simple combination of visual features and text descriptions in that the powerful perceptual and reasoning abilities of LMMs play a crucial role in discriminating between real and fake content.

\begin{table}[tb]
  \caption{Ablation analysis measured by AUC ($\%$). [Key: \textbf{Best}; Avg.: Average; Rec.: Diffusion Reconstruction Procedure; Fus. Dynamic Fusion Strategy].}
  \vspace{1mm}
  \label{tab:expt_ablation_study}
  \centering
    \resizebox{1.\linewidth}{!}{
  \begin{tabular}{ccccc|ccccccc|c}
    \hline
    \multicolumn{5}{c|}{Modules} & \multirow{2}{*}{{\makecell{Video-\\Crafter$1$}}} & \multirow{2}{*}{{\makecell{Zero\\scope}}} & \multirow{2}{*}{{\makecell{Open\\Sora}}} & \multirow{2}{*}{{\makecell{Sora}}} & \multirow{2}{*}{{\makecell{Pika}}} & \multirow{2}{*}{{\makecell{Stable \\ Diff.}}} & \multirow{2}{*}{{\makecell{Stable \\ Video}}} & \multirow{2}{*}{{\makecell{Avg.}}} \\
    ViT & Rec. & IAFA &MMFR & Fus. & & & & & & & &  \\
    \hline
    \rowcolor{gray!30}\cmark & & & & & $68.2 {\scriptstyle \,\pm\, 1.3}$ & $80.1 {\scriptstyle \,\pm\, 2.8}$ & $64.8 {\scriptstyle \,\pm\, 2.1}$ & $59.6 {\scriptstyle \,\pm\, 1.9}$ & $79.2 {\scriptstyle \,\pm\, 2.9}$ & $89.2 {\scriptstyle \,\pm\, 2.6}$ &$ 87.4 {\scriptstyle \,\pm\, 2.1}$ & $75.5 {\scriptstyle \,\pm\, 1.8}$\\
    \cmark & \cmark & & & &$70.1 {\scriptstyle \,\pm\, 1.6}$ & $76.3 {\scriptstyle \,\pm\, 1.9}$ &$77.6 {\scriptstyle \,\pm\, 1.8}$ & $66.2 {\scriptstyle \,\pm\, 2.0}$& $80.2 {\scriptstyle \,\pm\, 1.9}$ & $83.9 {\scriptstyle \,\pm\, 2.5}$ & $86.1 {\scriptstyle \,\pm\, 2.1}$ & $77.2 {\scriptstyle \,\pm\, 1.3}$ \\
    \rowcolor{gray!30}\cmark & \cmark & \cmark & & &$90.2 {\scriptstyle \,\pm\, 2.8}$ & $90.1 {\scriptstyle \,\pm\, 2.5}$ &$85.8 {\scriptstyle \,\pm\, 1.8}$ & $82.0 {\scriptstyle \,\pm\, 2.8}$ & $93.9 {\scriptstyle \,\pm\, 1.8}$ & $91.9 {\scriptstyle \,\pm\, 2.6}$ & $85.7 {\scriptstyle \,\pm\, 2.7}$ & $88.5 {\scriptstyle \,\pm\, 1.7}$ \\
    & & & \cmark & & $89.7 {\scriptstyle \,\pm\, 3.8}$ & $93.4 {\scriptstyle \,\pm\, 3.2}$ & $84.6 {\scriptstyle \,\pm\, 3.2}$ & $83.2 {\scriptstyle \,\pm\, 2.6}$ & $94.0 {\scriptstyle \,\pm\, 2.2}$ & $89.7 {\scriptstyle \,\pm\, 2.0}$ & $86.2 {\scriptstyle \,\pm\, 2.8}$ & $88.7 {\scriptstyle \,\pm\, 2.8}$\\
    \cmark & \cmark & \cmark & \cmark & &  $92.1 {\scriptstyle \,\pm\, 3.1}$ & $93.8 {\scriptstyle \,\pm\, 2.9}$ & $86.5 {\scriptstyle \,\pm\, 2.6}$ & $83.2 {\scriptstyle \,\pm\, 1.9}$ & $90.1 {\scriptstyle \,\pm\, 2.5}$ & $90.2 {\scriptstyle \,\pm\, 2.1}$ & $87.6 {\scriptstyle \,\pm\, 2.9}$ & $89.1 {\scriptstyle \,\pm\, 2.3}$\\
    \rowcolor{gray!30}\cmark & \cmark & \cmark & \cmark & \cmark &   $\mathbf{93.5} {\scriptstyle \,\pm\, 3.6}$ & $\mathbf{94.0} {\scriptstyle \,\pm\, 2.8}$ & $\mathbf{88.8} {\scriptstyle \,\pm\, 2.8}$ & $\mathbf{86.2} {\scriptstyle \,\pm\, 1.8}$ & $\mathbf{95.9} {\scriptstyle \,\pm\, 2.8}$ & $\mathbf{95.7} {\scriptstyle \,\pm\, 2.5}$ & $\mathbf{89.9} {\scriptstyle \,\pm\, 2.0}$ &  $\mathbf{92.0} {\scriptstyle \,\pm\, 2.6}$  \\
  \hline
  \end{tabular}
  }
\end{table}

\SubSection{Ablation Study}
Tab.~\ref{tab:expt_ablation_study} shows the impact of individual modules proposed in MM-Det.
Specifically, we use the Hybrid ViT~\cite{dosovitskiy2020image} as the base model and incorporate it with the reconstruction procedure for diffusion trace amplification, which enhances the detection performance by $+1.7\%$ AUC score. Such a module raises the performance in OpenSora and Sora, revealing that frequency-based information benefits forgery detection on these methods.
Detection performance is further increased by using IAFA, which strengthens the learning between in-frame and cross-frame information, increasing a $+13.0\%$ AUC score to the base model. The rise in performance indicates such temporal information benefits most types of forgery video detection.
After that, Tab.~\ref{tab:expt_ablation_study}'s line $4$ indicates the effectiveness of MMFR: a detector purely based on such representation can receive $88.7\%$ performance in AUC, $+13.2\%$ higher than the base model.
In addition, by merging MMFR (\textit{i.e.}, LMM branch) and ST branch, the performance rises by $+0.4\%$. 
Finally, with the dynamic fusion strategy, our method receives an impressive $92.0\%$ AUC score for all generative methods, higher than every single feature.
These experiments highlight the necessities of each module in our framework. Moreover, an ablation study on LLMs is detailed in Appendix~\ref{sec:supple_ablation_llms} to prove the effectiveness of various LLMs in MM-Det.

\begin{figure}[t]
    \centering
    \includegraphics[width=1.\textwidth]{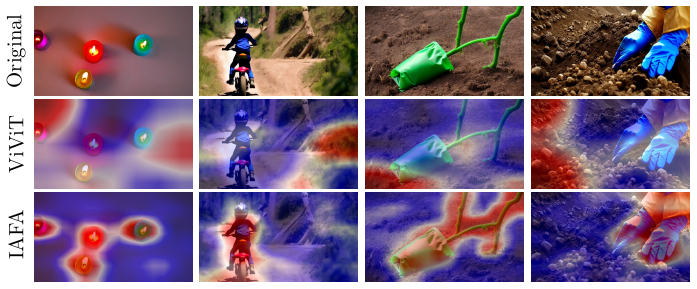}
    \caption{Visualization of artifacts captured from our IAFA and ViViT~\cite{arnab2021vivit}. We use activation maps to highlight spatial weights within each frame. All content is generated by VideoCrafter$1$. Features from the last layer of transformers are extracted for visualization.~\vspace{-3mm}
    }
    \label{fig:expt_st_heatmap}
\end{figure}

\SubSection{Spatio Temporal Information Anaylsis}
\label{sec:st_analysis}
In the analysis of IAFA, we visualize feature activation maps from the last layer of ViT in the ST branch based on the L2-norm. We compare our feature maps with another spatiotemporal baseline, ViViT~\cite{arnab2021vivit}, 
as depicted in Fig.~\ref{fig:expt_st_heatmap}. While attention maps of ViViT are sparse and irregular, the ones of our IAFA have a tendency to concentrate on the segmentation of diffusion-generated objects, indicating that IAFA captures typical spatial forgery regions in frames. The attention mainly focuses on common forgery traces, such as blurred generative patterns and defective parts of objects, signaling that diffusion models might find it difficult to generate delicate content. 
The concentration of activation on certain informative objects discloses both spatial artifacts of existing generative methods, demonstrating the effectiveness of our proposed ST branch.

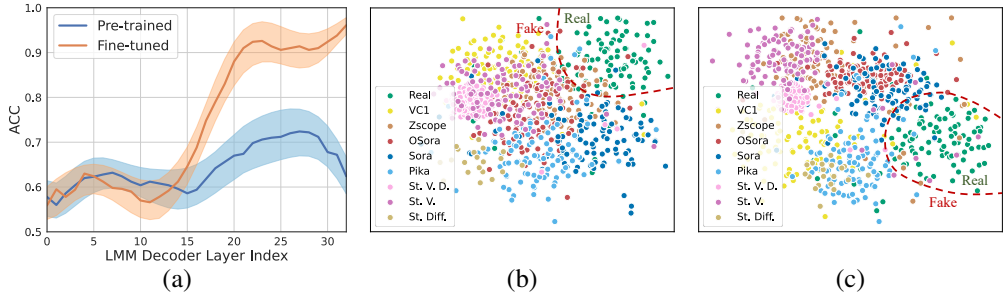
\begin{figure}[t]
    \centering
    \begin{tikzpicture}
        \begin{scope}[shift={(-3.6,0)}]
            \node at (0,0) {\includegraphics[width=0.335\textwidth]{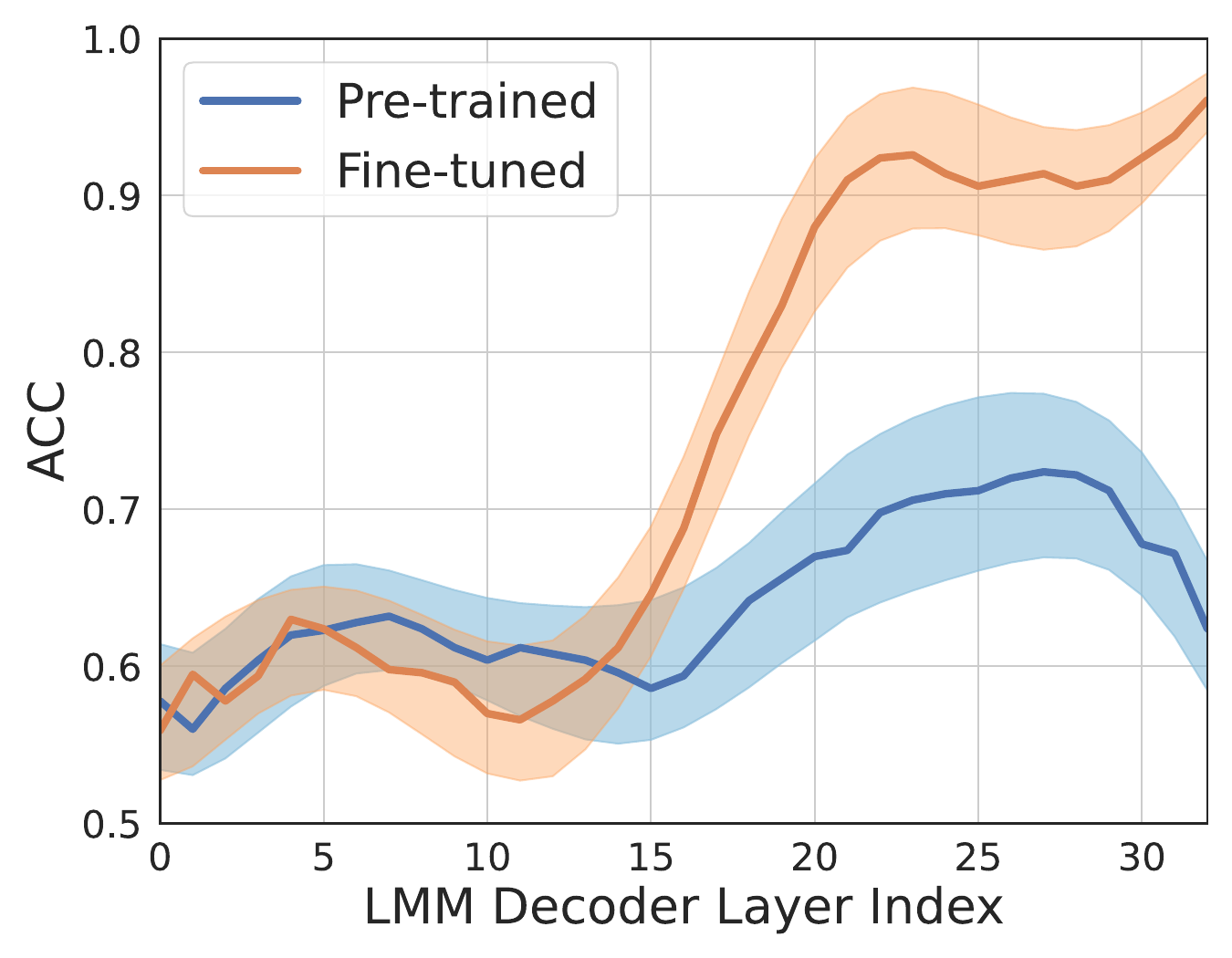}};
            \node at (0, -1.95) {(a)};
        \end{scope}
        \begin{scope}[shift={(0.987,0)}]
            \node at (0,0.182){\includegraphics[width=0.3037\textwidth]{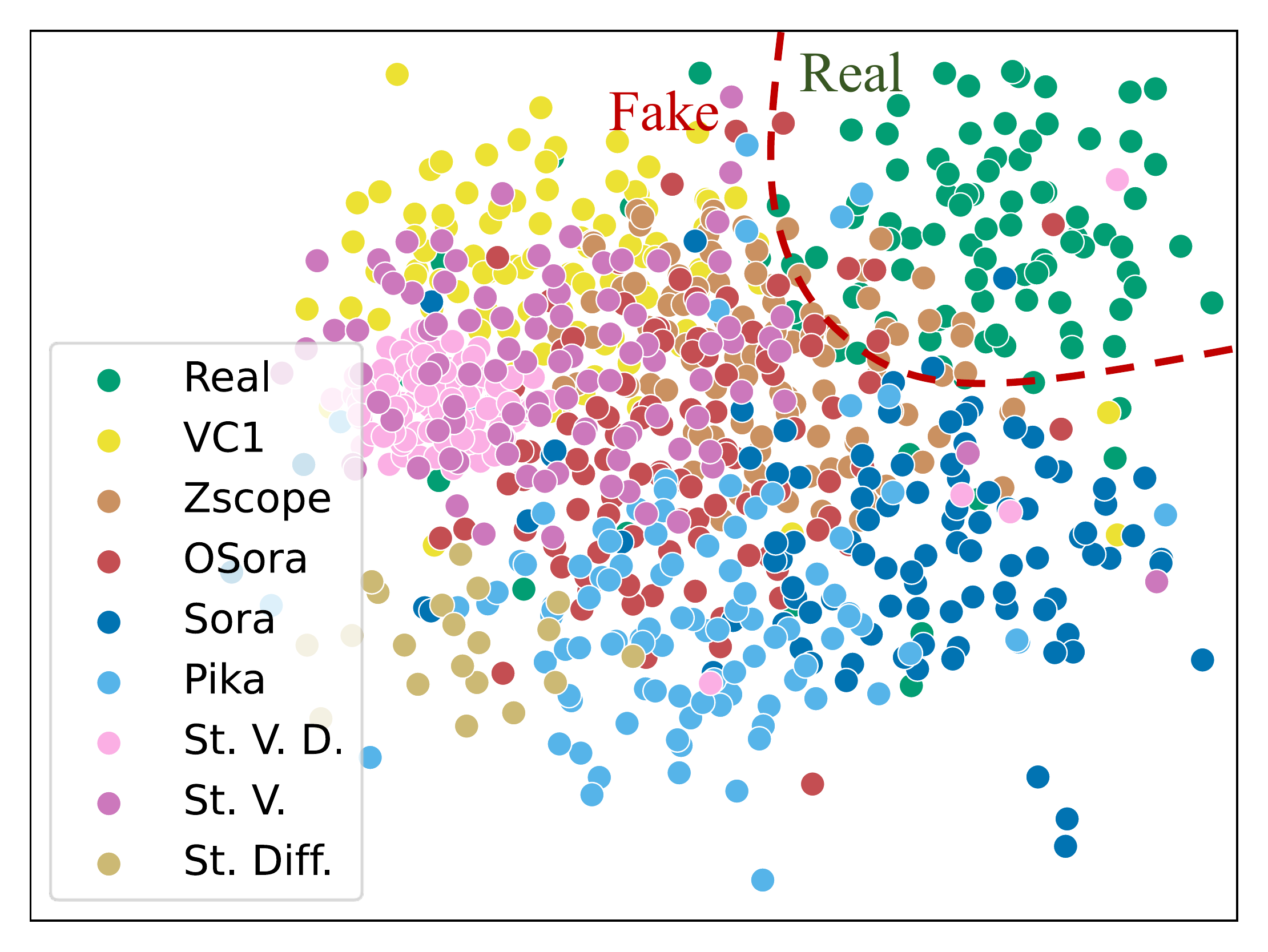}};
            \node at (0, -1.95) {(b)};
        \end{scope}
        \begin{scope}[shift={(5.35,0)}]
            \node at (0,0.182) {\includegraphics[width=0.3037\textwidth]{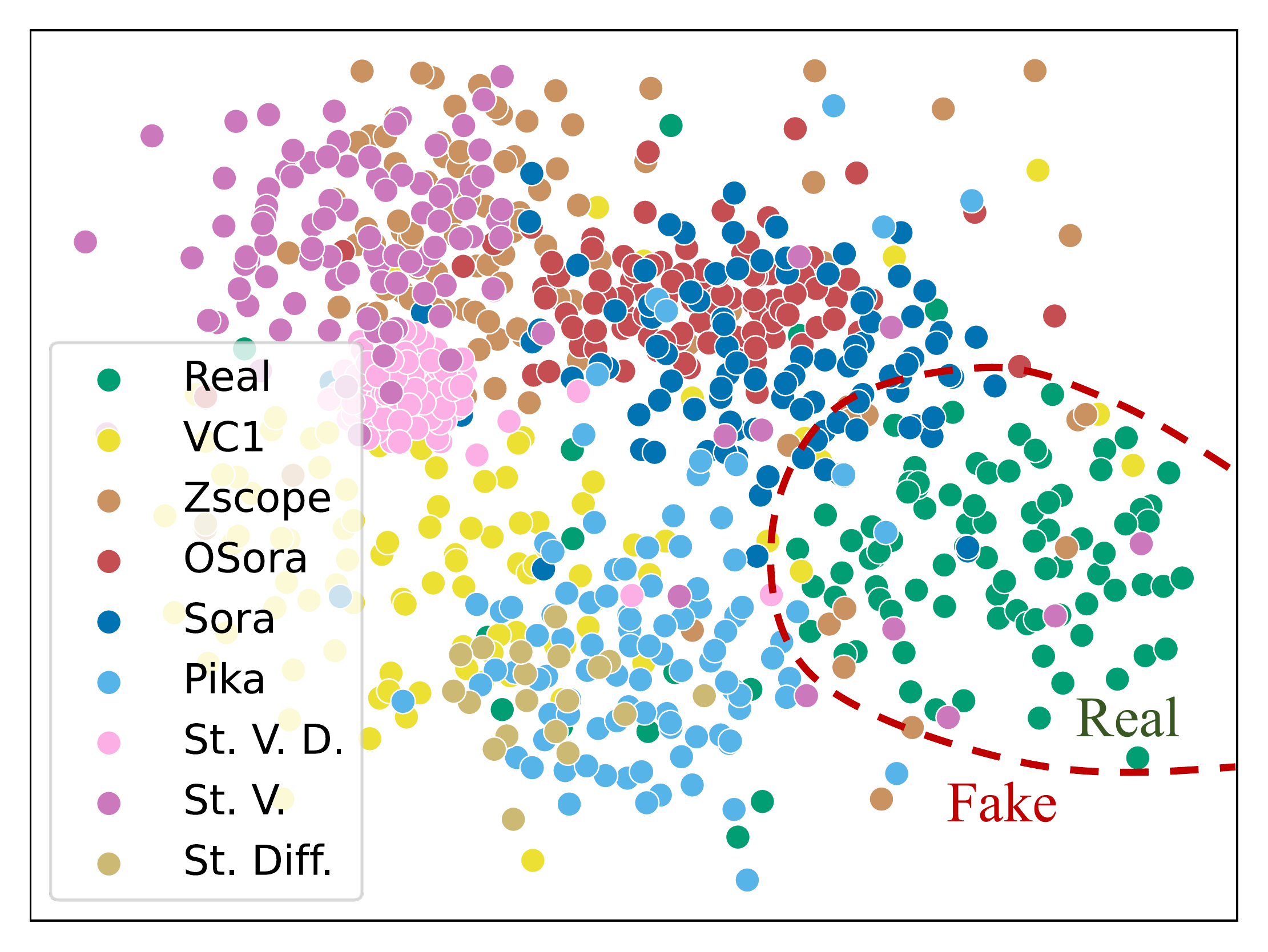}};
            \node at (0, -1.95) {(c)};
        \end{scope}
    
    \end{tikzpicture}
    \vspace{-1mm}
    \caption{(a):
    Clustering accuracy using features from different layers in LMM branch showcases MMFR's ability to discern forgeries. 
    (b)(c): t-SNE~\cite{van2008visualizing} visualization of features from ST and LMM branches. 
    For each dataset, $100$ videos are sampled and clustered for good visibility. Both features demonstrate boundaries between real and forgery videos.  
    (b) Features from the ST branch. (c): Features from the LMM branch.
    } \vspace{-5mm}
    \label{fig:expts_supple}
\end{figure}
\vspace{-1mm}

\SubSection{Multi-modal Forgery Representation Analysis}
\label{sec:mmfr_analysis}

Fig.~\ref{fig:expts_supple} details the effectiveness of MMFR in LMM branch.
First, as depicted in Fig.~\ref{fig:expts_supple}~\textcolor{red}{a}, we quantify the detection ability of features from each Transformer-based decoder layer in the LLM.
Specifically, we evaluate both pre-trained LLaVA~\cite{liu2024visual} and its fine-tuned version on the task of distinguishing real and fake frames. These frames are randomly sampled from $1,000$ real videos and equivalent fake ones from Stable Video Diffusion~\cite{rombach2022high}.
Layer-wise outputs from the large language model in LLaVA (\textit{i.e.}, Vicuna~\cite{touvron2023llama}) are obtained --- for $i$~th layer in the LLM, we denote its output features as $\mathbf{f}^{o}_i$, $i \in [1, 33]$.
The K-Means clustering algorithm is adopted to evaluate the classification accuracy based on $\mathbf{f}^{o}_i$. 
Empirically, we observe that features extracted from a fine-tuned LLaVA show promising classification accuracy in the last few layers, \textit{e.g.}, $22$-nd or later layers.
This phenomenon indicates that specific layers in LLMs indeed generate features that can be used for image forensic tasks. Such features are utilized in MMFR to exhibit high generalization ability towards diverse and unseen forgeries. 
Secondly, the comparison between pre-trained and fine-tuned LLaVA highlights the importance of downstream task-oriented instruction tuning for LMMs.
This conclusion is consistent with the findings of prior works~\cite{liu2023improved, liu2024visual, zhang2024llama, zhang2023video}.

In addition, shown in Fig.~\ref{fig:expts_supple}~\textcolor{red}{b} and ~\ref{fig:expts_supple}~\textcolor{red}{c}, we analyze features from ST branch and LMM branch through t-SNE~\cite{van2008visualizing}.
Both features achieve superior performance in separating real and forgery videos. In Fig.~\ref{fig:expts_supple}~\textcolor{red}{b}, spatiotemporal information forms a rough boundary between real and fake videos. This feature is effective for VideoCrafter$1$~\cite{chen2023videocrafter1}, Zeroscope, Stable Video, and Pika, whose durations and resolutions are similar to the training set. 
However, the detection performance might decrease on Sora and OpenSora with overlay in the clusters. We suppose that various resolutions and durations may compromise the generalization ability, magnifying the importance of a comprehensive dataset for these videos. Fig.~\ref{fig:expts_supple}~\textcolor{red}{c} demonstrates the more powerful feature from LMM branch. Samples from Zeroscope, Sora, and Pika are compacted into a denser area, indicating the ability of LLMs to conduct generalizable reasoning. Such features provide new insights for detection when spatial and temporal artifacts are not obvious among the latest forgery videos.

\vspace{-3mm}
\Section{Conclusion}
\vspace{-3mm}
In this work, we develop an effective video-level algorithm termed Multi-Modal Detection (MM-Det) for diffusion-generated video detection. 
MM-Det leverages a novel generalizable Multi-Modal Forgery Representation (MMFR) that is obtained from multi-modal spaces in LMMs.
Specifically, the proposed MM-Det has two major branches: the LMM branch, which incorporates vision and text features from the fine-tuned foundation model, and the ST branch, which concentrates on modeling spatial-temporal information aggregated through In-and-Across Frame Attention.
Extensive experiments demonstrate the effectiveness of our proposed detector. 
In addition, we establish a comprehensive dataset for various diffusion generative videos, which we hope will serve as a general benchmark for real-world video forensic tasks.

\Paragraph{Acknowledgement} This work was supported in part by the National Natural Science Foundation of China under Grant $62301310$ and $62225112$, and in part by Sichuan Science and Technology
Program under Grant $2024$NSFSC$1426$.

{
\small
\bibliographystyle{plain}
\bibliography{main}

\begin{thebibliography}{10}

\bibitem{abu2016youtube}
Sami Abu-El-Haija, Nisarg Kothari, Joonseok Lee, Paul Natsev, George Toderici, Balakrishnan Varadarajan, and Sudheendra Vijayanarasimhan.
\newblock Youtube-8m: A large-scale video classification benchmark.
\newblock {\em arXiv preprint arXiv:1609.08675}, 2016.

\bibitem{alayrac2022flamingo}
Jean-Baptiste Alayrac, Jeff Donahue, Pauline Luc, Antoine Miech, Iain Barr, Yana Hasson, Karel Lenc, Arthur Mensch, Katherine Millican, Malcolm Reynolds, et~al.
\newblock Flamingo: a visual language model for few-shot learning.
\newblock {\em Advances in neural information processing systems}, 35:23716--23736, 2022.

\bibitem{arnab2021vivit}
Anurag Arnab, Mostafa Dehghani, Georg Heigold, Chen Sun, Mario Lu{\v{c}}i{\'c}, and Cordelia Schmid.
\newblock Vivit: A video vision transformer.
\newblock In {\em Proceedings of the IEEE/CVF international conference on computer vision}, pages 6836--6846, 2021.

\bibitem{blattmann2023stable}
Andreas Blattmann, Tim Dockhorn, Sumith Kulal, Daniel Mendelevitch, Maciej Kilian, Dominik Lorenz, Yam Levi, Zion English, Vikram Voleti, Adam Letts, et~al.
\newblock Stable video diffusion: Scaling latent video diffusion models to large datasets.
\newblock {\em arXiv preprint arXiv:2311.15127}, 2023.

\bibitem{chen2023videocrafter1}
Haoxin Chen, Menghan Xia, Yingqing He, Yong Zhang, Xiaodong Cun, Shaoshu Yang, Jinbo Xing, Yaofang Liu, Qifeng Chen, Xintao Wang, et~al.
\newblock Videocrafter1: Open diffusion models for high-quality video generation.
\newblock {\em arXiv preprint arXiv:2310.19512}, 2023.

\bibitem{corvi2023intriguing}
Riccardo Corvi, Davide Cozzolino, Giovanni Poggi, Koki Nagano, and Luisa Verdoliva.
\newblock Intriguing properties of synthetic images: from generative adversarial networks to diffusion models.
\newblock In {\em Proceedings of the IEEE/CVF Conference on Computer Vision and Pattern Recognition}, pages 973--982, 2023.

\bibitem{corvi2023detection}
Riccardo Corvi, Davide Cozzolino, Giada Zingarini, Giovanni Poggi, Koki Nagano, and Luisa Verdoliva.
\newblock On the detection of synthetic images generated by diffusion models.
\newblock In {\em ICASSP 2023-2023 IEEE International Conference on Acoustics, Speech and Signal Processing (ICASSP)}, pages 1--5. IEEE, 2023.

\bibitem{cozzolino2023raising}
Davide Cozzolino, Giovanni Poggi, Riccardo Corvi, Matthias Nie{\ss}ner, and Luisa Verdoliva.
\newblock Raising the bar of ai-generated image detection with clip.
\newblock {\em arXiv preprint arXiv:2312.00195}, 2023.

\bibitem{cozzolino2024raising}
Davide Cozzolino, Giovanni Poggi, Riccardo Corvi, Matthias Nie{\ss}ner, and Luisa Verdoliva.
\newblock Raising the bar of ai-generated image detection with clip.
\newblock In {\em Proceedings of the IEEE/CVF Conference on Computer Vision and Pattern Recognition}, pages 4356--4366, 2024.

\bibitem{cozzolino2021id}
Davide Cozzolino, Andreas R{\"o}ssler, Justus Thies, Matthias Nie{\ss}ner, and Luisa Verdoliva.
\newblock Id-reveal: Identity-aware deepfake video detection.
\newblock In {\em Proceedings of the IEEE/CVF International Conference on Computer Vision}, pages 15108--15117, 2021.

\bibitem{dosovitskiy2020image}
Alexey Dosovitskiy, Lucas Beyer, Alexander Kolesnikov, Dirk Weissenborn, Xiaohua Zhai, Thomas Unterthiner, Mostafa Dehghani, Matthias Minderer, Georg Heigold, Sylvain Gelly, et~al.
\newblock An image is worth 16x16 words: Transformers for image recognition at scale.
\newblock {\em arXiv preprint arXiv:2010.11929}, 2020.

\bibitem{fu2024rawiw}
Kang Fu, Xiaohong Liu, Jun Jia, Zicheng Zhang, Yicong Peng, and Jia Wang.
\newblock Rawiw: Raw image watermarking robust to isp pipeline.
\newblock {\em Displays}, 82:102637, 2024.

\bibitem{gragnaniello2021gan}
Diego Gragnaniello, Davide Cozzolino, Francesco Marra, Giovanni Poggi, and Luisa Verdoliva.
\newblock Are gan generated images easy to detect? a critical analysis of the state-of-the-art.
\newblock In {\em 2021 IEEE international conference on multimedia and expo (ICME)}, pages 1--6. IEEE, 2021.

\bibitem{gu2021open}
Xiuye Gu, Tsung-Yi Lin, Weicheng Kuo, and Yin Cui.
\newblock Open-vocabulary object detection via vision and language knowledge distillation.
\newblock {\em arXiv preprint arXiv:2104.13921}, 2021.

\bibitem{gu2021spatiotemporal}
Zhihao Gu, Yang Chen, Taiping Yao, Shouhong Ding, Jilin Li, Feiyue Huang, and Lizhuang Ma.
\newblock Spatiotemporal inconsistency learning for deepfake video detection.
\newblock In {\em Proceedings of the 29th ACM international conference on multimedia}, pages 3473--3481, 2021.

\bibitem{xiao_neurips}
Xiao Guo, Vishal Asnani, Sijia Liu, and Xiaoming Liu.
\newblock Tracing hyperparameter dependencies for model parsing via learnable graph pooling network.
\newblock In {\em Proceeding of Thirty-eighth Conference on Neural Information Processing Systems}, Vancouver, Canada, December 2024.

\bibitem{xiao_hifi_net++}
Xiao Guo, Xiaohong Liu, Iacopo Masi, and Xiaoming Liu.
\newblock Language-guided hierarchical fine-grained image forgery detection and localization.
\newblock In {\em International Journal of Computer Vision}, December 2024.

\bibitem{guo2023hierarchical}
Xiao Guo, Xiaohong Liu, Zhiyuan Ren, Steven Grosz, Iacopo Masi, and Xiaoming Liu.
\newblock Hierarchical fine-grained image forgery detection and localization.
\newblock In {\em Proceedings of the IEEE/CVF Conference on Computer Vision and Pattern Recognition}, pages 3155--3165, 2023.

\bibitem{haliassos2021lips}
Alexandros Haliassos, Konstantinos Vougioukas, Stavros Petridis, and Maja Pantic.
\newblock Lips don't lie: A generalisable and robust approach to face forgery detection.
\newblock In {\em Proceedings of the IEEE/CVF conference on computer vision and pattern recognition}, pages 5039--5049, 2021.

\bibitem{he2016deep}
Kaiming He, Xiangyu Zhang, Shaoqing Ren, and Jian Sun.
\newblock Deep residual learning for image recognition.
\newblock In {\em Proceedings of the IEEE conference on computer vision and pattern recognition}, pages 770--778, 2016.

\bibitem{hu2021lora}
Edward~J Hu, Yelong Shen, Phillip Wallis, Zeyuan Allen-Zhu, Yuanzhi Li, Shean Wang, Lu~Wang, and Weizhu Chen.
\newblock Lora: Low-rank adaptation of large language models.
\newblock {\em arXiv preprint arXiv:2106.09685}, 2021.

\bibitem{jeong2022bihpf}
Yonghyun Jeong, Doyeon Kim, Seungjai Min, Seongho Joe, Youngjune Gwon, and Jongwon Choi.
\newblock Bihpf: Bilateral high-pass filters for robust deepfake detection.
\newblock In {\em Proceedings of the IEEE/CVF Winter Conference on Applications of Computer Vision}, pages 48--57, 2022.

\bibitem{li2023blip}
Junnan Li, Dongxu Li, Silvio Savarese, and Steven Hoi.
\newblock Blip-2: Bootstrapping language-image pre-training with frozen image encoders and large language models.
\newblock In {\em International conference on machine learning}, pages 19730--19742. PMLR, 2023.

\bibitem{li2022blip}
Junnan Li, Dongxu Li, Caiming Xiong, and Steven Hoi.
\newblock Blip: Bootstrapping language-image pre-training for unified vision-language understanding and generation.
\newblock In {\em International conference on machine learning}, pages 12888--12900. PMLR, 2022.

\bibitem{li2020face}
Lingzhi Li, Jianmin Bao, Ting Zhang, Hao Yang, Dong Chen, Fang Wen, and Baining Guo.
\newblock Face x-ray for more general face forgery detection.
\newblock In {\em Proceedings of the IEEE/CVF conference on computer vision and pattern recognition}, pages 5001--5010, 2020.

\bibitem{liu2024cfpl}
Ajian Liu, Shuai Xue, Jianwen Gan, Jun Wan, Yanyan Liang, Jiankang Deng, Sergio Escalera, and Zhen Lei.
\newblock Cfpl-fas: Class free prompt learning for generalizable face anti-spoofing.
\newblock {\em arXiv preprint arXiv:2403.14333}, 2024.

\bibitem{liu2023improved}
Haotian Liu, Chunyuan Li, Yuheng Li, and Yong~Jae Lee.
\newblock Improved baselines with visual instruction tuning.
\newblock {\em arXiv preprint arXiv:2310.03744}, 2023.

\bibitem{liu2024visual}
Haotian Liu, Chunyuan Li, Qingyang Wu, and Yong~Jae Lee.
\newblock Visual instruction tuning.
\newblock {\em Advances in neural information processing systems}, 36, 2024.

\bibitem{liu2024forgery}
Huan Liu, Zichang Tan, Chuangchuang Tan, Yunchao Wei, Jingdong Wang, and Yao Zhao.
\newblock Forgery-aware adaptive transformer for generalizable synthetic image detection.
\newblock In {\em Proceedings of the IEEE/CVF Conference on Computer Vision and Pattern Recognition}, pages 10770--10780, 2024.

\bibitem{liu2022pscc}
Xiaohong Liu, Yaojie Liu, Jun Chen, and Xiaoming Liu.
\newblock Pscc-net: Progressive spatio-channel correlation network for image manipulation detection and localization.
\newblock {\em IEEE Transactions on Circuits and Systems for Video Technology}, 32(11):7505--7517, 2022.

\bibitem{liu2022ts2}
Yuqi Liu, Pengfei Xiong, Luhui Xu, Shengming Cao, and Qin Jin.
\newblock Ts2-net: Token shift and selection transformer for text-video retrieval.
\newblock In {\em European conference on computer vision}, pages 319--335. Springer, 2022.

\bibitem{luo2024lare}
Yunpeng Luo, Junlong Du, Ke~Yan, and Shouhong Ding.
\newblock Lare\^{} 2: Latent reconstruction error based method for diffusion-generated image detection.
\newblock In {\em Proceedings of the IEEE/CVF Conference on Computer Vision and Pattern Recognition}, pages 17006--17015, 2024.

\bibitem{ma2023exposing}
Ruipeng Ma, Jinhao Duan, Fei Kong, Xiaoshuang Shi, and Kaidi Xu.
\newblock Exposing the fake: Effective diffusion-generated images detection.
\newblock {\em arXiv preprint arXiv:2307.06272}, 2023.

\bibitem{mandelli2022detecting}
Sara Mandelli, Nicol{\`o} Bonettini, Paolo Bestagini, and Stefano Tubaro.
\newblock Detecting gan-generated images by orthogonal training of multiple cnns.
\newblock In {\em 2022 IEEE International Conference on Image Processing (ICIP)}, pages 3091--3095. IEEE, 2022.

\bibitem{neimark2021video}
Daniel Neimark, Omri Bar, Maya Zohar, and Dotan Asselmann.
\newblock Video transformer network.
\newblock In {\em Proceedings of the IEEE/CVF international conference on computer vision}, pages 3163--3172, 2021.

\bibitem{ojha2023towards}
Utkarsh Ojha, Yuheng Li, and Yong~Jae Lee.
\newblock Towards universal fake image detectors that generalize across generative models.
\newblock In {\em Proceedings of the IEEE/CVF Conference on Computer Vision and Pattern Recognition}, pages 24480--24489, 2023.

\bibitem{pan2024towards}
Yongyang Pan, Xiaohong Liu, Siqi Luo, Yi~Xin, Xiao Guo, Xiaoming Liu, Xiongkuo Min, and Guangtao Zhai.
\newblock Towards effective user attribution for latent diffusion models via watermark-informed blending.
\newblock {\em arXiv preprint arXiv:2409.10958}, 2024.

\bibitem{qian2020thinking}
Yuyang Qian, Guojun Yin, Lu~Sheng, Zixuan Chen, and Jing Shao.
\newblock Thinking in frequency: Face forgery detection by mining frequency-aware clues.
\newblock In {\em European conference on computer vision}, pages 86--103. Springer, 2020.

\bibitem{radford2021learning}
Alec Radford, Jong~Wook Kim, Chris Hallacy, Aditya Ramesh, Gabriel Goh, Sandhini Agarwal, Girish Sastry, Amanda Askell, Pamela Mishkin, Jack Clark, et~al.
\newblock Learning transferable visual models from natural language supervision.
\newblock In {\em International conference on machine learning}, pages 8748--8763. PMLR, 2021.

\bibitem{ricker2022towards}
Jonas Ricker, Simon Damm, Thorsten Holz, and Asja Fischer.
\newblock Towards the detection of diffusion model deepfakes.
\newblock {\em arXiv preprint arXiv:2210.14571}, 2022.

\bibitem{ricker2024aeroblade}
Jonas Ricker, Denis Lukovnikov, and Asja Fischer.
\newblock Aeroblade: Training-free detection of latent diffusion images using autoencoder reconstruction error.
\newblock {\em arXiv preprint arXiv:2401.17879}, 2024.

\bibitem{rombach2022high}
Robin Rombach, Andreas Blattmann, Dominik Lorenz, Patrick Esser, and Bj{\"o}rn Ommer.
\newblock High-resolution image synthesis with latent diffusion models.
\newblock In {\em Proceedings of the IEEE/CVF conference on computer vision and pattern recognition}, pages 10684--10695, 2022.

\bibitem{sha2023fake}
Zeyang Sha, Zheng Li, Ning Yu, and Yang Zhang.
\newblock De-fake: Detection and attribution of fake images generated by text-to-image generation models.
\newblock In {\em Proceedings of the 2023 ACM SIGSAC Conference on Computer and Communications Security}, pages 3418--3432, 2023.

\bibitem{song2020denoising}
Jiaming Song, Chenlin Meng, and Stefano Ermon.
\newblock Denoising diffusion implicit models.
\newblock {\em arXiv preprint arXiv:2010.02502}, 2020.

\bibitem{su2023pandagpt}
Yixuan Su, Tian Lan, Huayang Li, Jialu Xu, Yan Wang, and Deng Cai.
\newblock Pandagpt: One model to instruction-follow them all.
\newblock {\em arXiv preprint arXiv:2305.16355}, 2023.

\bibitem{tan2024rethinking}
Chuangchuang Tan, Yao Zhao, Shikui Wei, Guanghua Gu, Ping Liu, and Yunchao Wei.
\newblock Rethinking the up-sampling operations in cnn-based generative network for generalizable deepfake detection.
\newblock In {\em Proceedings of the IEEE/CVF Conference on Computer Vision and Pattern Recognition}, pages 28130--28139, 2024.

\bibitem{tan2023learning}
Chuangchuang Tan, Yao Zhao, Shikui Wei, Guanghua Gu, and Yunchao Wei.
\newblock Learning on gradients: Generalized artifacts representation for gan-generated images detection.
\newblock In {\em Proceedings of the IEEE/CVF Conference on Computer Vision and Pattern Recognition}, pages 12105--12114, 2023.

\bibitem{team2023gemini}
Gemini Team, Rohan Anil, Sebastian Borgeaud, Yonghui Wu, Jean-Baptiste Alayrac, Jiahui Yu, Radu Soricut, Johan Schalkwyk, Andrew~M Dai, Anja Hauth, et~al.
\newblock Gemini: a family of highly capable multimodal models.
\newblock {\em arXiv preprint arXiv:2312.11805}, 2023.

\bibitem{thies2016face2face}
Justus Thies, Michael Zollhofer, Marc Stamminger, Christian Theobalt, and Matthias Nie{\ss}ner.
\newblock Face2face: Real-time face capture and reenactment of rgb videos.
\newblock In {\em Proceedings of the IEEE conference on computer vision and pattern recognition}, pages 2387--2395, 2016.

\bibitem{touvron2023llama}
Hugo Touvron, Louis Martin, Kevin Stone, Peter Albert, Amjad Almahairi, Yasmine Babaei, Nikolay Bashlykov, Soumya Batra, Prajjwal Bhargava, Shruti Bhosale, et~al.
\newblock Llama 2: Open foundation and fine-tuned chat models.
\newblock {\em arXiv preprint arXiv:2307.09288}, 2023.

\bibitem{van2017neural}
Aaron Van Den~Oord, Oriol Vinyals, et~al.
\newblock Neural discrete representation learning.
\newblock {\em Advances in neural information processing systems}, 30, 2017.

\bibitem{van2008visualizing}
Laurens Van~der Maaten and Geoffrey Hinton.
\newblock Visualizing data using t-sne.
\newblock {\em Journal of machine learning research}, 9(11), 2008.

\bibitem{wang2020cnn}
Sheng-Yu Wang, Oliver Wang, Richard Zhang, Andrew Owens, and Alexei~A Efros.
\newblock Cnn-generated images are surprisingly easy to spot... for now.
\newblock In {\em Proceedings of the IEEE/CVF conference on computer vision and pattern recognition}, pages 8695--8704, 2020.

\bibitem{wang2023internvid}
Yi~Wang, Yinan He, Yizhuo Li, Kunchang Li, Jiashuo Yu, Xin Ma, Xinhao Li, Guo Chen, Xinyuan Chen, Yaohui Wang, et~al.
\newblock Internvid: A large-scale video-text dataset for multimodal understanding and generation.
\newblock {\em arXiv preprint arXiv:2307.06942}, 2023.

\bibitem{wang2023dire}
Zhendong Wang, Jianmin Bao, Wengang Zhou, Weilun Wang, Hezhen Hu, Hong Chen, and Houqiang Li.
\newblock Dire for diffusion-generated image detection.
\newblock In {\em Proceedings of the IEEE/CVF International Conference on Computer Vision}, pages 22445--22455, 2023.

\bibitem{wang2023altfreezing}
Zhendong Wang, Jianmin Bao, Wengang Zhou, Weilun Wang, and Houqiang Li.
\newblock Altfreezing for more general video face forgery detection.
\newblock In {\em Proceedings of the IEEE/CVF Conference on Computer Vision and Pattern Recognition}, pages 4129--4138, 2023.

\bibitem{wodajo2021deepfake}
Deressa Wodajo and Solomon Atnafu.
\newblock Deepfake video detection using convolutional vision transformer.
\newblock {\em arXiv preprint arXiv:2102.11126}, 2021.

\bibitem{wu2023cheap}
Guangyang Wu, Weijie Wu, Xiaohong Liu, Kele Xu, Tianjiao Wan, and Wenyi Wang.
\newblock Cheap-fake detection with llm using prompt engineering.
\newblock In {\em 2023 IEEE International Conference on Multimedia and Expo Workshops (ICMEW)}, pages 105--109. IEEE, 2023.

\bibitem{wu2023next}
Shengqiong Wu, Hao Fei, Leigang Qu, Wei Ji, and Tat-Seng Chua.
\newblock Next-gpt: Any-to-any multimodal llm.
\newblock {\em arXiv preprint arXiv:2309.05519}, 2023.

\bibitem{wu2019mantra}
Yue Wu, Wael AbdAlmageed, and Premkumar Natarajan.
\newblock Mantra-net: Manipulation tracing network for detection and localization of image forgeries with anomalous features.
\newblock In {\em Proceedings of the IEEE/CVF conference on computer vision and pattern recognition}, pages 9543--9552, 2019.

\bibitem{xing2023dynamicrafter}
Jinbo Xing, Menghan Xia, Yong Zhang, Haoxin Chen, Xintao Wang, Tien-Tsin Wong, and Ying Shan.
\newblock Dynamicrafter: Animating open-domain images with video diffusion priors.
\newblock {\em arXiv preprint arXiv:2310.12190}, 2023.

\bibitem{xu2023tall}
Yuting Xu, Jian Liang, Gengyun Jia, Ziming Yang, Yanhao Zhang, and Ran He.
\newblock Tall: Thumbnail layout for deepfake video detection.
\newblock In {\em Proceedings of the IEEE/CVF international conference on computer vision}, pages 22658--22668, 2023.

\bibitem{yan2024transcending}
Zhiyuan Yan, Yuhao Luo, Siwei Lyu, Qingshan Liu, and Baoyuan Wu.
\newblock Transcending forgery specificity with latent space augmentation for generalizable deepfake detection.
\newblock In {\em Proceedings of the IEEE/CVF Conference on Computer Vision and Pattern Recognition}, pages 8984--8994, 2024.

\bibitem{yang2024diffstega}
Yiwei Yang, Zheyuan Liu, Jun Jia, Zhongpai Gao, Yunhao Li, Wei Sun, Xiaohong Liu, and Guangtao Zhai.
\newblock Diffstega: Towards universal training-free coverless image steganography with diffusion models.
\newblock {\em arXiv preprint arXiv:2407.10459}, 2024.

\bibitem{yao2024reverse}
Yuguang Yao, Xiao Guo, Vishal Asnani, Yifan Gong, Jiancheng Liu, Xue Lin, Xiaoming Liu, Sijia Liu, et~al.
\newblock Reverse engineering of deceptions on machine-and human-centric attacks.
\newblock {\em Foundations and Trends{\textregistered} in Privacy and Security}, 6(2):53--152, 2024.

\bibitem{yu2019attributing}
Ning Yu, Larry~S Davis, and Mario Fritz.
\newblock Attributing fake images to gans: Learning and analyzing gan fingerprints.
\newblock In {\em Proceedings of the IEEE/CVF international conference on computer vision}, pages 7556--7566, 2019.

\bibitem{zhang2023next}
Ao~Zhang, Liming Zhao, Chen-Wei Xie, Yun Zheng, Wei Ji, and Tat-Seng Chua.
\newblock Next-chat: An lmm for chat, detection and segmentation.
\newblock {\em arXiv preprint arXiv:2311.04498}, 2023.

\bibitem{zhang2023video}
Hang Zhang, Xin Li, and Lidong Bing.
\newblock Video-llama: An instruction-tuned audio-visual language model for video understanding.
\newblock {\em arXiv preprint arXiv:2306.02858}, 2023.

\bibitem{zhang2024llama}
Renrui Zhang, Jiaming Han, Chris Liu, Aojun Zhou, Pan Lu, Yu~Qiao, Hongsheng Li, and Peng Gao.
\newblock Llama-adapter: Efficient fine-tuning of large language models with zero-initialized attention.
\newblock In {\em The Twelfth International Conference on Learning Representations}, 2024.

\bibitem{zhang2024common}
Yue Zhang, Ben Colman, Xiao Guo, Ali Shahriyari, and Gaurav Bharaj.
\newblock Common sense reasoning for deepfake detection.
\newblock In {\em European Conference on Computer Vision}, 2025.

\bibitem{zhang2024vision}
Yue Zhang, Ziqiao Ma, Jialu Li, Yanyuan Qiao, Zun Wang, Joyce Chai, Qi~Wu, Mohit Bansal, and Parisa Kordjamshidi.
\newblock Vision-and-language navigation today and tomorrow: A survey in the era of foundation models.
\newblock {\em arXiv preprint arXiv:2407.07035}, 2024.

\bibitem{zhang2024spartun3d}
Yue Zhang, Zhiyang Xu, Ying Shen, Parisa Kordjamshidi, and Lifu Huang.
\newblock Spartun3d: Situated spatial understanding of 3d world in large language models.
\newblock {\em arXiv preprint arXiv:2410.03878}, 2024.

\bibitem{zhao2021multi}
Hanqing Zhao, Wenbo Zhou, Dongdong Chen, Tianyi Wei, Weiming Zhang, and Nenghai Yu.
\newblock Multi-attentional deepfake detection.
\newblock In {\em Proceedings of the IEEE/CVF conference on computer vision and pattern recognition}, pages 2185--2194, 2021.

\bibitem{zheng2021exploring}
Yinglin Zheng, Jianmin Bao, Dong Chen, Ming Zeng, and Fang Wen.
\newblock Exploring temporal coherence for more general video face forgery detection.
\newblock In {\em Proceedings of the IEEE/CVF international conference on computer vision}, pages 15044--15054, 2021.

\bibitem{zhu2023minigpt}
Deyao Zhu, Jun Chen, Xiaoqian Shen, Xiang Li, and Mohamed Elhoseiny.
\newblock Minigpt-4: Enhancing vision-language understanding with advanced large language models.
\newblock {\em arXiv preprint arXiv:2304.10592}, 2023.

\end{thebibliography}
}



\section{Appendix / Supplemental Material}
\subsection{Limitations}
Although our proposed MM-Det advances in detecting fake videos, further issues are left to handle. First, as the landscape of video manipulation technology evolves, new techniques and tools will outpace existing detection methods. The gap between training data and real-world applications can lead to misleading results. Currently, our method is limited to fully synthesized diffusion videos, lacking the generalization to more delicate forgeries like partial manipulation. A possible reason is that small forgery traces disappear after multiple downsampling operations in the deep network of LMMs. Besides, the integration of a large language model into detection costs huge computational complexity, which is not an optimal choice for an environment with limited resources.

In conclusion, while our algorithm makes a critical step forward in detecting fake videos, it faces significant challenges due to the rapid advancement of video manipulation technologies. Addressing these limitations requires further research to keep pace with the evolving techniques in digital content manipulation.

\subsection{Broader Impacts}
In this work, our team has developed an effective algorithm to detect fake videos, a breakthrough that promises to fortify the authenticity of online media. In real-world social media where misinformation can spread rapidly, our method acts as a crucial safeguard by empowering platforms to flag and remove deceptive videos before they can mislead users. However, our methods may fail in extreme situations, such as blurred images or noisy images. The algorithm should be carefully treated to avoid misleading results. The long-term influence of our work protects public trust by ensuring the authenticity of digital content. Meanwhile, precaution is needed for a fair application.

\begin{table}[b]
\renewcommand*{\thetable}{S1}
  \caption{Diffusion Video Forensics Composition [Key: T$2$V: Text-to-Video; I$2$V: Image-to-Video]}
  \label{tab:supple_dvf_dataset_intro}
  \centering
  \begin{tabular}{cccc}
    \toprule
    Dataset & Source & Video Number & Resolution  \\
    \midrule
    Real & Youtube \& Internvid-$10$M & $2,750$ & $1,280 \times 720$ \\
    Stable Video Diffusion & I$2$V & $1,800$ & $1,024 \times 576$\\
    VideoCrafter$1$ & T$2$V & $450$ &$1,024 \times 576$\\
    Zeroscope & T$2$V & $800$ &$1,024 \times 576$\\
    Sora & Social Media & $153$ & $1,280 \times 720$\\
    Pika & T$2$V, I$2$V & $122$ & $1,280 \times 720$\\
    OpenSora & T$2$V & $500$ &$512 \times 512$\\
    Stable Diffusion & I$2$V & $12$ &$1,080 \times 1,920$\\
    Stable Video & T$2$V & $101$ & $1,024 \times 576$, $1,920 \times 1,080$\\

  \bottomrule
  \end{tabular}
\end{table}
\subsection{Datasets}
\subsubsection{Diffusion Video Forensics}
\label{sec:supple_dvf}
 We propose a comprehensive dataset, named Diffusion Video Forensics (DVF), for diffusion video forensics, as shown in Tab. \ref{tab:supple_dvf_dataset_intro}. 
DVF consists of fake videos generated from $8$ different generation methods, covering text-to-video and image-to-video generative methods. In total, We make a collection of $2,788$ real videos and $4,111$ fake videos. Real videos are from YouTube and Internvid-$10$M~\cite{wang2023internvid}.

We formally introduce the generation pipeline for video collection. Generation methods in DVF are divided into closed-sourced methods and open-sourced methods. For closed-sourced methods(Sora, Pika, Stable Diffusion~\cite{rombach2022high} and Stable Video), we collect video samples from official websites and social media like TikTok to form the forgery video datasets. For open-sourced methods, the generation pipelines are divided into text-to-video (OpenSora, VideoCrafter1~\cite{chen2023videocrafter1} and Zeroscope) and image-to-video (Stable Video Diffusion~\cite{blattmann2023stable}). For a text-to-video generative method, real data derives from a text-image paired video dataset, Internvid-$10$M. Specifically, we fetch paired real videos $R = \{\mathbf{r}_{1}, \mathbf{r}_{2}, ..., \mathbf{r}_{N}\}$ and corresponding captions $C = \{c_{1}, c_{2}, ..., c_{N}\}$. We directly apply the captions as prompts to generate fake video datasets $F$, such that $F = \{\mathbf{f}_{1}, \mathbf{f}_{2}, ... \mathbf{f}_{N}\},\mathbf{f}_{i} = g_t(c_{i}), i \in [1, N]$, where $g_t$ denotes a text-to-video method. Both $R$ and $F$ are included in DVF as real and fake datasets. For image-to-video methods, real videos come from Youtube-$8$M~\cite{abu2016youtube}, which are denoted as $R = \{\mathbf{r}_{1}, \mathbf{r}_{2}, ..., \mathbf{r}_{N}\}, \mathbf{r}_i \in \mathbb{R}^{L \times H \times W \times C}$. For each video $r_i$, a real frame $x_i \in \mathbb{R}^{H \times W \times C}$ is randomly sampled from $r_i$ and serves as the conditional input for generation. The fake datasets are obtained as $F = \{\mathbf{f}_{1}, \mathbf{f}_{2}, ... \mathbf{f}_{N}\},\mathbf{f}_{i} = g_{im}(r_{i}), i \in [1, N]$, where $g_{im}$ denotes an image-to-video method.

\begin{figure}[t]
\renewcommand*{\thefigure}{S1}
\centering
\includegraphics[width=1.\linewidth]{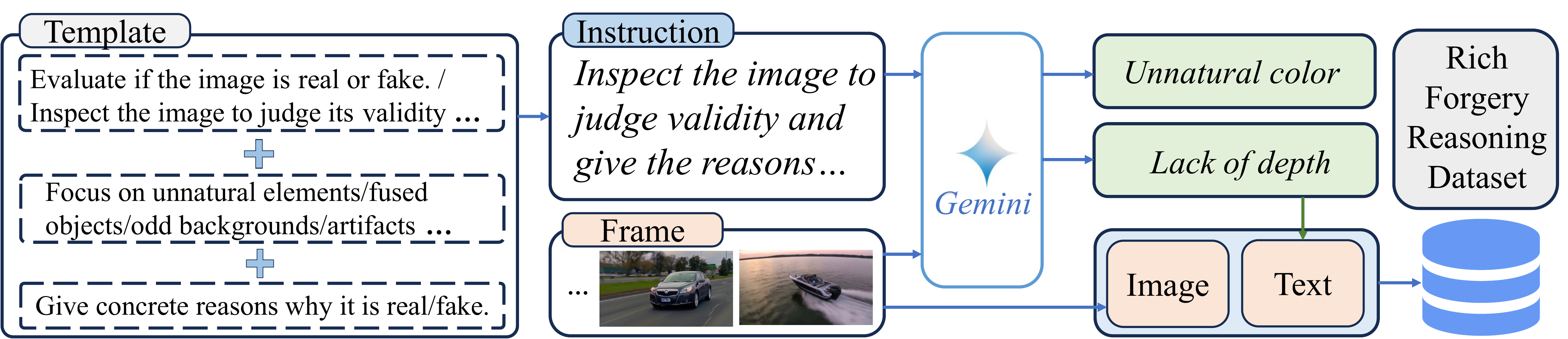}
\caption{The overview of Rich Forgery Reasoning Dataset. To obtain text-image paired data on forgery reasoning, we take advantage of the powerful reasoning ability of Gemini~\cite{team2023gemini} to generate ground truth for each image. First, an instruction $T$ is sampled from a template, along with a frame $\mathbf{f} \in \mathbb{R}^{H \times W \times C}$ fed into Gemini for forgery analysis and detection. The response $r$ contains detailed judgment and reasoning on the content and authenticity of $\mathbf{f}$(\textit{e.g.}, analyses on color and depth). Finally, $\mathbf{f}$ and $r$ are collected as the text-image paired data to form Rich Forgery Reasoning Dataset.}
\label{fig:supple_rfrd}
\end{figure}

\subsubsection{Rich Forgery Reasoning Dataset}
\label{sec:supple_rfrd}
We construct a text-image paired dataset, called Rich Forgery Reasoning Dataset (RFRD), to support instruction-tuning LMMs on the forgery detection task, as shown in Fig.~\ref{fig:supple_rfrd}. We start from the YouTube and Stable Video Diffusion dataset in DVF, where we select $1,000$ real videos and $1,800$ fake videos. Depending on the powerful reasoning ability of Gemini~\cite{team2023gemini} v$1.5$ Pro, we follow the scheme in Fig.~\ref{fig:supple_rfrd} to generate ground truth for frames. In total, $1,921$ real frames and $3,579$ fake frames are sampled to generate $5,500$ image-text paired textual descriptions. These descriptions are then cleaned and converted into $38$k multi-turn conversations for fine-tuning LLaVA~\cite{liu2024visual} in LMM branch of MM-Det.

\subsection{Implementation Details}
\label{sec:supple_implementation}
\paragraph{Hyperparameters of MM-Det}

We introduce the implementation of our MM-Det. In ST branch, we employ a Hybrid-ViT~\cite{dosovitskiy2020image} to build the video-level feature encoder. We choose Hybrid-ViT-B, with a ViT-B/$16$ on top of a ResNet-$50$ backbone, the patch size $14 \times 14$, and the hidden size $768$. We employ IAFA in each attention block of the ViT, with FC-tokens initialized as the class token of ViT. In the embedding stage, learnable spatial and temporal embeddings are introduced in the form of addition. All patches at the same position within frames share the same spatial embedding and patches with the same timestep share the same time embedding. A pre-trained VQ-VAE is applied to reconstruct videos, with hidden size $d = 256$ and the codebook size $K = 512$. We train the VQ-VAE on $50,000$ images from ImageNet. In LMM branch, we utilize LLaVA~\cite{liu2024visual} v$1.5$, with a CLIP~\cite{radford2021learning} encoder $\mathcal{E}$ of CLIP-ViT-L-patch$14$-$336$ and a large language model $\mathcal{D}$ of Vicuna-$7$b for reasoning. Our proposed MMFR is composed of the pooler output $\mathbf{F}_{V} \in \mathbb{R}^{1024}$ from the last layer of $\mathcal{E}$ and the embedding of output $\mathbf{F}_L$ from the last layer of $\mathcal{D}$. To balance effectiveness and efficiency, we fix the length of output tokens into $64$ and obtain $\mathbf{F}_L \in \mathbb{R}^{64 \times 4096}$.

\paragraph{Training and Inference}

As for the experimental resources in training and inference, we conduct all experiments using a single NVIDIA RTX $3090$ GPU and a maximum of $200$G memory. 

The training strategy of MM-Det is in two-stage. We first conduct instruction tuning for LLaVA in the LMM branch based on LoRA~\cite{hu2021lora}. We start from a pre-trained LLaVA v$1.5$ and train it on our collected Rich Forgery Reasoning Dataset detailed in Sec. ~\ref{sec:supple_rfrd}. We use an Adam optimizer with the learning rate set as $2e^{-5}$ for $10$ epochs. After that, we integrate LLaVA into MM-Det and conduct the overall training. The training set is split into $9:1$ for training and validation data. For each video, successive $10$ frames are randomly sampled and cropped into $224 \times 224$ as the input. We use an Adam optimizer with the learning rate set as $1e^{-4}$ for training until the model converges.

For inference, we evaluate all models at the video level. For frame-level baselines, the score of an entire video is obtained as the average score of all frames. For video-level methods, successive clips are fed according to the corresponding window size, and the entire score is obtained as the average score of all clips.
In addition, to leverage the efficiency for inference, MM-Det first caches MMFR for each video by conducting reasoning at the interval of $1$ frame every $6$ seconds. During inference, each video clip directly applies MMFR from the nearest cached frame as an approximation to reduce the huge computational cost. For each baseline, we conduct evaluations on $5$ different seeds ($1, 100, 999, 1234, 9999$) and choose the average score.

\begin{table}[t]
  \renewcommand*{\thetable}{S2}
  \caption{Evaluation on different Large Language Models measured by AUC(\%). [Key: \textbf{Best}; Avg.: Average].}
  \label{tab:supple_ablation_llms}
  \centering
  \resizebox{0.98\linewidth}{!}{

  \begin{tabular}{c|ccccccc|c}
    \hline
    \multirow{2}{*}{{\makecell{LLM}}} & \multirow{2}{*}{{\makecell{Video-\\Crafter$1$}}}  & \multirow{2}{*}{{\makecell{Zeroscope}}} & \multirow{2}{*}{{\makecell{OpenSora}}} & \multirow{2}{*}{{\makecell{Sora}}} & \multirow{2}{*}{{\makecell{Pika}}} & \multirow{2}{*}{{\makecell{Stable \\ Diffusion}}} & \multirow{2}{*}{{\makecell{Stable \\ Video}}} & \multirow{2}{*}{{\makecell{Avg.}}} \\
     & & & & & & & &  \\
    \hline
     N/A & $90.2 {\scriptstyle \,\pm\, 2.8}$ & $90.1 {\scriptstyle \,\pm\, 2.5}$ &$85.8 {\scriptstyle \,\pm\, 1.8}$ & $82.0 {\scriptstyle \,\pm\, 2.8}$ & $93.9 {\scriptstyle \,\pm\, 1.8}$ & $91.9 {\scriptstyle \,\pm\, 2.6}$ & $85.7 {\scriptstyle \,\pm\, 2.7}$ & $88.5 {\scriptstyle \,\pm\, 1.7}$\\
     Vicuna-$7$b & $93.5 {\scriptstyle \,\pm\, 3.6}$ & $\mathbf{94.0} {\scriptstyle \,\pm\, 2.8}$ & $88.8 {\scriptstyle \,\pm\, 2.8}$ & $\mathbf{86.2} {\scriptstyle \,\pm\, 1.8}$ & $95.9 {\scriptstyle \,\pm\, 2.8}$ & $95.7 {\scriptstyle \,\pm\, 2.5}$ & $89.9 {\scriptstyle \,\pm\, 2.0}$  &  $92.0  {\scriptstyle \,\pm\, 2.6}$ \\
     Vicuna-$13$b & $\mathbf{95.5} {\scriptstyle \,\pm\, 2.5} $ & $93.2 {\scriptstyle \,\pm\, 2.2}$  & $\mathbf{89.2} {\scriptstyle \,\pm\, 3.1}$ & $83.9 {\scriptstyle \,\pm\, 2.5}$ & $\mathbf{96.6} {\scriptstyle \,\pm\, 1.9}$ & $\mathbf{95.9} {\scriptstyle \,\pm\, 2.3}$ & $\mathbf{90.8} {\scriptstyle \,\pm\, 2.7}$ & $\mathbf{92.2} {\scriptstyle \,\pm\, 2.4}$ \\
     Mistral-$7$b & $92.6 {\scriptstyle \,\pm\, 2.9}$ & $92.1 {\scriptstyle \,\pm\, 2.5}$ & $86.3 {\scriptstyle \,\pm\, 3.6}$ & $83.2 {\scriptstyle \,\pm\, 2.2}$ & $94.6 {\scriptstyle \,\pm\, 2.6}$ & $93.6 {\scriptstyle \,\pm\, 2.0}$ & $86.2 {\scriptstyle \,\pm\, 2.8}$ & $89.8 {\scriptstyle \,\pm\, 2.6}$ \\

  \hline
  \end{tabular}
  }
\end{table}

\subsection{Ablation Study on LLMs}
\label{sec:supple_ablation_llms}
We adopt alternative LLMs in our MM-Det to evaluate different choices of language backbones. Performance is reported in Tab.~\ref{tab:supple_ablation_llms}.
More formally, for a fair comparison, when using different LLMs, we maintain other components \textit{e.g.}, CLIP, the reconstruction procedure, IAFA, and the dynamic fusion of the original MM-Det remained. 
Specifically, the introduction of a combined vision and text space from Vicuna-$7$b in LLaVA improves the performance by $+3.5\%$. 
As for the choice of LMMs, Vicuna-$7$b achieves an average AUC score of $92.0\%$, $+2.2\%$ higher than Mistral-$7$b. We suppose this result may be attributed to different attention mechanisms in Vicuna and Mistral. Vicuna-$13$b gains a further improvement by $+0.2\%$ due to incremental parameters in capturing more effective multi-modal feature spaces.
These results prove that our MMFR is effective and extensible to other language models.

\begin{table}[t]
  \renewcommand*{\thetable}{S3}
  \caption{Perfomance of MM-Det on common post-processing operations measured by AUC(\%).}

  \centering
  \resizebox{0.85\linewidth}{!}{

  \begin{tabular}{cccccc}
    \hline
    N/A & Blur $\sigma=3$ & JPEG $Q=50$ & Resize $0.7$ & Rotate $90$ & Mixed \\
    \hline
    $92.0$ & $86.2$ & $91.1$ & $89.9$ & $90.1$ & $88.6$ \\
    \hline
  \end{tabular}
  }
  \label{tab:supple_robustness}
\end{table}

\subsection{Robustness Analysis}
\label{sec:supple_robustness}
To analyze the robustness of our method, we conduct an additional evaluation of MM-Det based on common post-processing operations. We choose Gaussian blur with $\sigma = 3$, JPEG compression with quality $Q = 90$, resize with a ratio of $0.7$, rotation with an angle of $90$, and a mixture of all operations as unseen perturbations in real-world scenarios. Testing samples are selected from DVF to form a total of $500$ real videos and $500$ fake videos. As reported in Tab.~\ref{tab:supple_robustness}, MM-Det meets a degradation of $0.9\%$(JPEG Compression) to $5.8\%$(Gaussian blur), with all performance above $86\%$. The results indicate the effectiveness of our method under these operations. 
\end{document}